\documentclass[manuscript, screen]{acmart}

\AtBeginDocument{%
  }

\setcopyright{cc}
\setcctype{by}
\acmJournal{CSUR}
\acmYear{2026} \acmVolume{1} \acmNumber{1} \acmArticle{}
\acmMonth{1} \acmDOI{10.1145/3800686}

\acmISBN{978-1-4503-XXXX-X/18/06}

\usepackage{tikz}
\usepackage{longtable}
\usepackage{multirow}
\usepackage{ctable} 
\usepackage{natbib} 

\usepackage[]{xcolor}
\usepackage[]{hyperref}
\hypersetup{pdfborder=0 0 0}
\usepackage{subcaption}
\usepackage[normalem]{ulem}
\usepackage{adjustbox}

\usepackage{siunitx}
\usepackage{tcolorbox}

\def\finalmanuscript{1} 

\ifx\undefined\finalmanuscript
\newcommand{\CL}[1]{\textcolor{blue}{#1}}
\newcommand{\CLstrike}[1]{\textcolor{blue}{\sout{#1}}}
\newcommand{\ds}[1]{\textcolor{blue}{#1}}
\newcommand{\dsstrike}[1]{\textcolor{blue}{\sout{#1}}}

\else
\newcommand{\CL}[1]{\textcolor{black}{#1}}
\newcommand{\CLstrike}[1]{} 
\newcommand{\ds}[1]{\textcolor{black}{#1}}
\newcommand{\dsstrike}[1]{}
\fi

\newcounter{rqcounter}  
\renewcommand{\therqcounter}{RQ\arabic{rqcounter}} 

\newcommand{\RQdef}[2]{         
    \refstepcounter{rqcounter}  
    \therqcounter: #1           
    \label{rq:#2}               
}
\newcommand{\RQ}[1]{\hypersetup{linkcolor=black}~\ref{rq:#1}\hypersetup{linkcolor=violet}}
\newcommand{\RQnospace}[1]{\hypersetup{linkcolor=black}\ref{rq:#1}\hypersetup{linkcolor=violet}}
\newcommand{\RQinColor}[1]{(\ref{rq:#1})} 

\begin{document}


\title{FPGA-Enabled Machine Learning Applications in Earth Observation: A Systematic Review}

\author{Cédric Léonard}
\email{cedric.leonard@dlr.de}
\orcid{0009-0004-2252-2173}
\affiliation{%
  \institution{Technical University of Munich}
  \city{Munich}
  \country{Germany}
}
\affiliation{%
  \institution{Remote Sensing Technology Institute (IMF), German Aerospace Center (DLR)}
  \city{Weßling}
  \country{Germany}
}

\author{Dirk Stober}
\email{dirk.stober@tum.de}
\orcid{0009-0001-5096-7063}
\affiliation{%
  \institution{Technical University of Munich}
  \city{Munich}
  \country{Germany}
}

\author{Martin Schulz}
\email{schulzm@in.tum.de}
\orcid{0000-0001-9013-435X}
\affiliation{%
  \institution{Technical University of Munich}
  \city{Munich}
  \country{Germany}
}


\begin{abstract}
New UAV technologies and the NewSpace era are transforming Earth Observation missions and data acquisition. 
Numerous small platforms generate large data volume, straining bandwidth and requiring onboard decision-making to transmit high-quality information in time.
While Machine Learning allows real-time autonomous processing, FPGAs balance performance with adaptability to mission-specific requirements, enabling onboard deployment.
This review systematically analyzes \CL{$68$}\CLstrike{$66$} experiments deploying ML models on FPGAs for Remote Sensing applications.
We introduce two distinct taxonomies to capture both efficient model architectures and FPGA implementation strategies.
For transparency and reproducibility, we follow PRISMA 2020 guidelines and share all data and code at \url{https://github.com/CedricLeon/Survey_RS-ML-FPGA}. 
\end{abstract}


\received{20 February 2007}
\received[revised]{12 March 2009}
\received[accepted]{5 June 2009}

\begin{teaserfigure}
    \includegraphics[trim={0.1cm 12.6cm 4.7cm 4.8cm},clip,width=\textwidth]{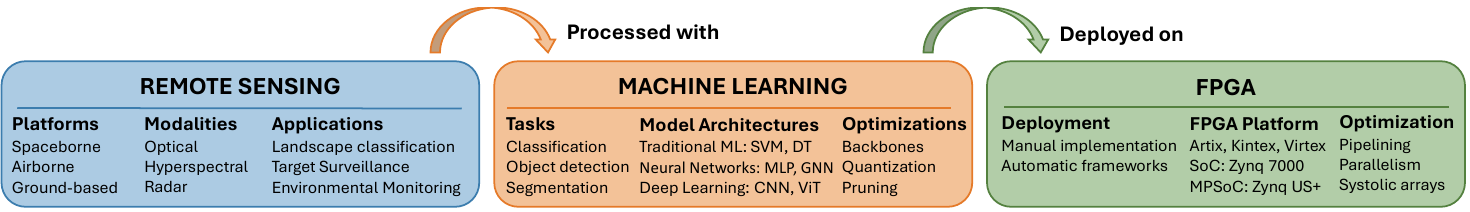}
    \Description[]{}
    \caption{\CL{Scope and content of the survey.}}
    \label{fig:conceptual_diagram}
\end{teaserfigure}

\maketitle

\newpage
\section{Introduction} 

The last decade of research has seen major innovations in Machine Learning (ML), particularly in Deep Learning (DL). Every discipline has seen publications leveraging ML's abstract learning capabilities to alleviate specific problems of the field.
Nowadays, many state-of-the-art solutions incorporate ML methods in their pipeline.
For instance, automatic navigation, for cars and Unmanned Aerial Vehicles (UAVs) alike, heavily relies on scene segmentation and obstacle detection. 
Similarly, video and image encoding apply learnable heuristics to select optimal patterns and splits in the data.
Beyond vision, language processing benefits from the wide context available through transformer tokenization, significantly enhancing performance.
By and large, these AI technologies have significant computational costs, making them inadequate in resource-constrained environments.
Therefore, many applications deploying their algorithms close to the sensor, e.g., cars and UAVs On-Board Computers (OBCs), GoPros, or intelligent microphones, benefit from lightweight optimizations of these computationally intensive models.
\CLstrike{Approximate computing optimizations, trading computation load over accuracy, such as quantization, can typically reduce memory overhead by a factor of $4$ and computational costs by a factor of $16$ \cite{nagelWhitePaperNeural2021}.}  

In this survey, we take a systematic look at one application of ML for computer vision: Remote Sensing (RS) for Earth Observation (EO)\footnote{
\CL{Throughout this survey, we will prefer the term RS, as it is the core technology of EO, which also encompasses in-situ measurements.}
\CLstrike{Throughout this survey we will prefer the term Remote Sensing (RS) rather than Earth Observation (EO).
RS specifically refers to acquiring information about the Earth's surface from a distance while EO includes both remote sensing and in situ measurements.}
}.
Contemporary challenges, such as the climate crisis, urban planning, or defense concerns, transformed RS into an essential field.
In particular, the rising NewSpace\footnote{\url{https://philab.esa.int/tag/new-space/}} industry has decreased the costs and the production time of SmallSats.
Because deploying CubeSats has never been so easy, we can observe a steady increase in the number of missions---to date, over 2600 nanosatellites have been launched \cite{kimOnOrbitAICloud2024}. 
However, payload restrictions on these compact satellites reduce their onboard compute capacity relative to their larger counterparts, forcing them to run application-specific routines in flight and underscoring the need for efficient onboard data processing.
In the particular example of RS, while the quantity of data grows, the bandwidth resources for the data transmission (downlink) stagnate \cite{pitonakCloudSatNet1FPGABasedHardwareAccelerated2022}.
Because of such limitations, most current missions acquire data on-demand or over certain pre-configured conditions, inevitably missing out relevant information.
The increasing sensing resolution and number of in-orbit satellites, paired with further restrictions on the radio-frequency spectrum, will worsen the problem \cite{ruzickaRaVAEnUnsupervisedChange2022}.
Processing the data directly onboard, transmitting only compressed or application-tailored information, constitutes an elegant solution for future missions.
Equipping future missions with edge computing platforms opens up possibilities to relieve data processing stress.
Because SmallSats' standard payloads typically include FPGA-based hardware \cite{siegleMitigationRadiationEffects2015, georgeOnboardProcessingHybrid2018}, this survey focuses on analyzing the literature deploying ML-based solutions for RS applications on FPGAs \ds{(Fig.~\ref{fig:conceptual_diagram})}. \\
The main contributions of this survey are:
\begin{itemize}
    \item \textbf{Systematic literature collection and analysis} ensuring a rigorous and transparent review process following the PRISMA 2020 guidelines 
    \CL{(Section \ref{section:methodology})}.
    \item \textbf{Comprehensive evaluation} of existing works, structured around eight Research Questions (RQs) to analyze the research landscape \CL{(Section \ref{section:landscape}), the design strategies (Section \ref{section:design})}, and method synergies \CL{(Section \ref{section:onboard})}.
    \item \textbf{Two specialized taxonomies}, one structuring RS applications based on their ML problem formulation \CL{(Section \ref{section:landscape})}, and another categorizing FPGA implementations by frameworks and design patterns \CL{(Section \ref{section:design})}.
    \item \textbf{Identification of research gaps}, structured under five subtopics to guide future advancements in FPGA-enabled ML for Earth Observation \CL{(Section~\ref{section:discussion})}. 
    \item \textbf{Guidelines and recommendations} to facilitate benchmarking, enhance reproducibility, and drive future research efforts \CL{(Section~\ref{section:discussion})}. 
\end{itemize}



\subsection{Background} 

Remote Sensing (RS) is the science of obtaining reliable information about objects or areas on the Earth's surface without direct physical contact \cite{campbellIntroductionRemoteSensing2011}.
It involves the acquisition, processing, and interpretation of data captured by sensors that detect electromagnetic radiation. 
The discipline is broadly divided into active and passive approaches.
Active systems, such as radar and LiDAR, emit their own energy signals and record the returning echoes, while passive systems depend on external sources of energy, primarily sunlight.
Furthermore, RS platforms are classified by their operational altitude into spaceborne and airborne systems.
Spaceborne platforms, typically mounted on satellites, provide extensive temporal and spatial coverage.
They enable continuous monitoring over vast areas and across diverse time scales---from near real-time updates to the analysis of long-term environmental trends \cite{zhuDeepLearningRemote2017}.
Airborne platforms, on the other hand, offer higher spatial resolution over more localized areas, making them ideal for detailed studies that require fine-grained spatial information.

%
Machine Learning (ML) methods have existed for several decades, but the past ten years have seen an unprecedented surge in their capabilities, largely driven by advances in computing power.
On one hand, Deep Learning (DL) abilities to explore and learn from high-dimensional spaces have precipitated the growth of the discipline.
In particular, foundation models and their generalization powers have been an intense research area over the last two years \cite{szwarcmanPrithviEO20VersatileMultiTemporal2025}.
On the other hand, traditional ML methods such as Decision Trees (DT) \cite{quinlanInductionDecisionTrees1986} or Support Vector Machines (SVM) \cite{cortesSupportvectorNetworks1995} are small and explainable methods still preferred in certain scenarios.
Nevertheless, the vast majority of current ML research focuses on DL, equally in RS, where Convolutional Neural Network (CNN) and Vision Transformer (ViT) abilities to learn from images lay a solid basis for many RS applications \cite{liuConvNet2020s2022a, szwarcmanPrithviEO20VersatileMultiTemporal2025}.

CPUs and, to a lesser extent, GPUs support a wide range of algorithms and kernels, requiring general-purpose hardware units such as multi-level caches, prefetching, and branch prediction.
Many specific algorithms do not fully utilize these features, leading to wasted energy and transistors.
FPGAs, by contrast, are reconfigurable integrated circuits adaptable to a specific algorithm, instantiating only the necessary hardware units.
FPGAs' reconfigurability comes at the cost of lower operating frequency, requiring comparatively higher parallelism to compete with CPUs and the already highly parallel GPUs.
In addition, such a high degree of freedom requires careful development.
Programming FPGAs is similar to designing an integrated circuit and is traditionally done at low-level using Hardware Description Languages (HDL).
However, with the growing adoption of AI on FPGAs, higher-level end-to-end toolchains are now available to implement Neural Networks (NNs) with popular kernels, such as CNNs \cite{AMDVitisAI,umurogluFINNFrameworkFast2017}.

\subsection{Related Work} 
\newcommand{\cmark}{\textcolor{green}{\checkmark}}
\newcommand{\smark}{\textcolor{orange}{$\approx$}}
\newcommand{\xmark}{\textcolor{red}{$\times$}}
\begin{table}
    \centering
    \newcolumntype{H}{>{\setbox0=\hbox\bgroup}c<{\egroup}@{}}
    \caption{Summary of related surveys. \cmark ~ covered, \xmark ~ not covered, \smark ~ partially covered.}
    \label{tab:related_surveys}
    \begin{tabular}{lcHccccc}
        \textbf{Reference} & \textbf{Year} & \textbf{Survey} & \textbf{RS Focus} & \textbf{ML Focus} & \textbf{FPGAs} & \textbf{Onboard} & \textbf{Years analyzed}\\
        \toprule
        \citet{lopezPromiseReconfigurableComputing2013} & 2013  & \cmark & \cmark       & \xmark    & \cmark    & Space     &     --     \\
        \citet{bouhaliFPGAApplicationsUnmanned2017} & 2017      & \cmark & \smark       & \xmark    & \cmark             & UAV       &     --     \\
        \citet{zhuDeepLearningRemote2017} & 2017                & \cmark & \cmark       & DL        & \xmark           & \xmark    & 2014--2017 \\
        \citet{abdelouahabAcceleratingCNNInference2018} & 2018  & \cmark & \xmark       & CNNs      & \cmark             & \xmark    & 2015--2018 \\
        \citet{georgeOnboardProcessingHybrid2018} & 2018        & \cmark & \smark       & \xmark    & \cmark    & Space     & 2000--2016 \\
        \citet{shawahnaFPGABasedAcceleratorsDeep2019} & 2019    & \cmark & \xmark       & DL        & \cmark             & \xmark    & 2009--2018 \\
        \citet{oscoReviewDeepLearning2021} & 2021               & \cmark & \cmark       & DL        & \xmark           & UAV       & 2012--2020 \\
        \citet{langeMachineLearningSpace2024} & 2024            & \cmark & \smark       & DL        & \xmark           & Space     & 2018--2023 \\
        \textbf{Our survey} & 2025                                       & \cmark & \cmark       & DL + ML   & \cmark    & \cmark    & 2014--2024  \\
        \bottomrule 
    \end{tabular}
\end{table}

We conduct this survey to combine the overlapping disciplines of Remote Sensing, Machine Learning, and FPGA deployment.
To the authors' knowledge, no work has yet explored the intersection of these three research domains. 
However, multiple reviews have covered the usage of ML in RS, explored FPGAs for onboard processing, or investigated FPGA-based accelerators for ML separately---see Table~\ref{tab:related_surveys} for some relevant literature reviews.
Notably, two surveys investigate the overlap between DL and FPGA design: \citet{abdelouahabAcceleratingCNNInference2018} and \citet{shawahnaFPGABasedAcceleratorsDeep2019}.
Both studies draft a comprehensive overview of design paradigms to implement DL methods on FPGAs.
Taking a closer look at onboard environments, \citet{georgeOnboardProcessingHybrid2018} and \citet{bouhaliFPGAApplicationsUnmanned2017} depict the specificities of onboard flying platforms, space- and airborne, respectively.
With a similar scope, \citet{langeMachineLearningSpace2024} specifically focus on the impact and mitigation techniques of radiation effects on DL models.
\section{Methodology}\label{section:methodology}
This study follows the Preferred Reporting Items for Systematic Reviews and Meta-Analyses (PRISMA) 2020 guidelines \cite{pagePRISMA2020Explanation2021} applicable to Computer Science, ensuring a complete, transparent, and reproducible review process.

\subsection{Research Questions}

This survey aims to examine articles using ML methods on FPGA hardware and discussing RS in EO.
We specifically focus on RS sensors mounted on aerial vehicles (drones or UAVs) or spaceborne satellites.
We explore this research intersection to summarize the community's work towards implementing automated methods onboard flying platforms, space- or airborne.
To provide a clear structure for this survey, we address the following eight Research Questions (RQs).
First, in Section~\ref{section:landscape}, RQ1-3 establish the research landscape and define the population of studies included in our analysis.
\begin{itemize}
    \item \RQdef{Which Remote Sensing applications are addressed, and how are they formulated as ML problems?}{applications}
    \item \RQdef{Which Machine Learning models are most prevalent, and what are their key architectural characteristics?}{models} 
    \item \RQdef{What are the motivations for choosing FPGAs over other computing platforms (CPUs, GPUs, ASICs)?}{FPGA-motivations} 
\end{itemize}
Then in Section~\ref{section:design}, RQ4-6 explore the technical details of the deployed solutions. Here, we catalog ML optimization methods, FPGA design paradigms, and various levels of parallelism available to designers.
\begin{itemize}
    \item \RQdef{What are the effective optimization strategies to reduce model footprint and computational complexity?}{optimizations} 
    \item \RQdef{How do FPGA design frameworks and patterns affect development effort and final performance?}{FPGA-framework} 
    \item \RQdef{How do different FPGA design approaches, such as pipelining, parallelization, and memory optimization, balance performance and constraints?}{FPGA-designs} 
\end{itemize}
Finally in Section~\ref{section:onboard}, R7-8 take a holistic view of the accomplishments in this field, discussing key achievements and, in line with the current NewSpace era, the potential of AI on the edge for \CL{RS}\CLstrike{EO} missions.
\begin{itemize}
    \item \RQdef{\CL{Given limited FPGA resources, which ML algorithms are suitable, and which RS applications can this platform support?}\CLstrike{What combination of Machine Learning model, optimization strategies, and FPGA design best synergizes to optimize key Remote Sensing metrics?}}{synergies} 
    \item \RQdef{How are AI-powered edge computing solutions transforming \CL{RS}\CLstrike{EO} missions?}{onboard} 
\end{itemize}

 
To answer these research questions, we conduct a systematic search procedure adapted from the method proposed by the PRISMA 2020 guidelines \cite{pagePRISMA2020Explanation2021}.
The procedure starts by defining a prompt to frame for the population of records of interest.
After collecting the results from the chosen research databases, we perform an initial screening pass via metadata analysis to filter out undesired records.
The second screening pass involves reviewing the complete records, excluding any not clearly identified as being outside the survey's scope based on the metadata alone. \CL{The final curated collection of studies is enriched with additional articles coming from private libraries.}

\subsection{Search Procedure}
We systematically retrieve relevant studies by querying Web of Science\footnote{\url{https://www.webofscience.com/}, last consulted on 06 March 2025} with the prompt\footnote{The "\textbf{TOPICS}" field searches across each item's title, abstract, author keywords, and keyword plus in the database. The prompt is case-insensitive.} below:

\begin{centering}
    \small
    \textit{\textbf{TOPICS}:  \textbf{\big{[}} (Machine Learning \textbf{OR} Deep Learning \textbf{OR} Neural Network \textbf{OR} Convolutional Neural Network \textbf{OR} Recurrent Neural Network \textbf{OR} ML \textbf{OR} DL \textbf{OR} NN \textbf{OR} CNN \textbf{OR} RNN)
    \textbf{AND} \\
    (Earth Observation \textbf{OR} Remote Sensing \textbf{OR} Earth Science \textbf{OR} LiDAR \textbf{OR} SAR \textbf{OR} UAV \textbf{OR} Sentinel)
    \textbf{AND} \\
    (FPGA \textbf{OR} Field-Programmable Gate Array \textbf{OR} Field Programmable Gate Array \textbf{OR} Versal) \big{]}\\
    \textbf{YEAR PUBLISHED}: $2014$-$2024$, 
    \textbf{LANGUAGE}: ENGLISH \\}
\end{centering}

The central part of the prompt consists of three expressions framing for any ML model, the usage of RS data in EO, and the explicit mention of FPGA devices.
We choose not to consider articles published more than 10 years ago.
Indeed, the FPGA technology, the state-of-the-art DL models, and even the tackled problems have evolved rapidly.
Therefore, work from more than 10 years ago is often too different from modern studies to be compared fairly. 
After retrieving the $119$ results of the prompt, we proceed to screen the articles to filter mismatches. 
To this effect, we pick 4 Exclusion Criteria (EC):
\begin{itemize}
    \item EC1 - Discussion and review papers, without experiments.
    \item EC2 - Publication unrelated to or not using ML models;
    \item EC3 - Publication unrelated to or not using RS data for EO;
    \item EC4 - Publication unrelated to or not using FPGA devices;
\end{itemize}
While EC1 and EC2 are self-explanatory, EC3 mostly rules out non-EO applications of remote sensing technology, e.g., LiDAR for autonomous driving or SAR calibration.
EC4 excludes studies with no experiments or details about the FPGA implementation.
For example, EC4 eliminates records mentioning FPGA hardware as a possible solution or using a different FPGA acronym, e.g., fast patch-free global learning.
We design this filtering process to retain only the most relevant, experiment-based studies that reflect the current state of FPGA-based ML in Remote Sensing.
After screening, $46$ articles are kept for review \CL{for a total of $48$ records when adding the records from private libraries}. 
Each article included in the survey is thoroughly read and tagged in Zotero \cite{Zotero2006}. 
In particular, we use Zotero's tagging system and API to systematically report the experiments' specificities and performance metrics. 
The entire database extracted via Zotero's API, along with the code used for analysis, figures, and main taxonomy tables, is available on GitHub at \url{https://github.com/CedricLeon/Survey_RS-ML-FPGA}.
Our repository also hosts supplemental material, such as extended methodological considerations \CL{and the PRISMA 2020 flow diagram}.

\section{Mapping the Research Landscape: Applications, Models, and Hardware}\label{section:landscape}
In this section, we analyze the motivations, tackled problems, and solutions presented by each surveyed study.
\CLstrike{In particular, we propose a taxonomy based on the ML and RS components of the experiments in Table~\ref{table:rs-ml_taxonomy}.}


\begin{table}
\centering

\newcolumntype{H}{>{\setbox0=\hbox\bgroup}c<{\egroup}@{}}

\caption{RS/ML Taxonomy Table}
\label{table:rs-ml_taxonomy}

\begin{adjustbox}{totalheight=\textheight-2\baselineskip}
\begin{tabular}{cccccccccHcccc}

\multicolumn{1}{c}{\textbf{ML}} & \multicolumn{3}{c}{\textbf{Remote Sensing}} & \multicolumn{2}{c}{\textbf{Article}} & \multicolumn{4}{c}{\textbf{ML Model}} & \multicolumn{1}{c}{\textbf{FPGA}} &\multicolumn{3}{c}{\textbf{Performance}}\\
\cmidrule(lr){1-1} \cmidrule(lr){2-4} \cmidrule(lr){5-6} \cmidrule(lr){7-10} \cmidrule(lr){11-11} \cmidrule(lr){12-14}
\textbf{Task} &\textbf{Mod.} &\textbf{Application} &\textbf{Dataset} &\textbf{Ref.} &\textbf{Year} &\textbf{Original Name} &\textbf{Core} &\textbf{Backbone} &\textbf{Eq. Model} &\textbf{Family} &\textbf{Score [\%]} &\textbf{MF[MB]} &\textbf{C[OP]}\\
 \toprule 
\multirow{24}{*}{\rotatebox[origin=c]{90}{Classification \hspace{16mm}}}
   &\multirow{15}{*}{RGB}
       &\multirow{6}{*}{Cloud coverage class.}
           &\multirow{2}{*}{L8 Biome \cite{fogaCloudDetectionAlgorithm2017}}
               &\multirow{1}{*}{\cite{pitonakCloudSatNet1FPGABasedHardwareAccelerated2022}}
                   &\multirow{1}{*}{2022}
                       &\multirow{1}{*}{CloudSatNet-1 Q2}
                           &\multirow{1}{*}{Custom CNN}
                               &\multirow{1}{*}{-}
                                   &\multirow{1}{*}{CNN}
                                       &\multirow{1}{*}{Zynq 7000}
                                           &\multirow{1}{*}{83.41 (OA)}
                                               &\multirow{1}{*}{1.43}
                                                   &\multirow{1}{*}{-}\\
\end{tabular}
\end{adjustbox}
\end{table}

\subsection{Remote Sensing/Machine Learning Taxonomy}\label{section:ml-rs_taxonomy}
\CL{To help readers explore studies addressing similar problem settings, we present in Table~\ref{table:rs-ml_taxonomy} a taxonomy grouping the collected studies based on how RS applications are formulated as ML tasks.}
\CLstrike{To help readers explore studies addressing similar problem settings, we group the collected studies by how RS problems are formulated as ML tasks.
Table~\ref{table:rs-ml_taxonomy} presents this taxonomy, clustering all reported experiments across the selected studies.}
When a study conducts several relevant experiments, we extract each corresponding model and metric.
For example, \citet{suhAlgorithmHardwareCoOptimizationEnergyEfficient2021} present a methodological contribution and evaluate several sizes of the same architecture, i.e., SSD 0.25x, SSD 0.5x, and SSD 1.0x.
We argue that reporting each experiment separately, rather than considering a single experiment per study, portrays a more complete picture of the conducted work.
In total, the taxonomy includes \CL{$68$}\CLstrike{$66$} experiments drawn from \CL{$48$}\CLstrike{$46$} studies.

Each row in Table~\ref{table:rs-ml_taxonomy} corresponds to an experiment.
\CLstrike{The first group of columns, dedicated to the \textit{RS Problem}, introduces the taxonomy: experiments are hierarchically grouped by ML \textit{’Task’}, i.e., the family of the ML problem, data modality (\textit{’Mod.’}), i.e., the type of RS data used as input, \textit{’Application’} and \textit{’Dataset’}.}\ds{In the first column, we classify the ML problem into different Computer Vision (CV) tasks. To facilitate comparison with other ML works, we describe the ML \textit{’Task’} using CV terminology over RS nomenclature\footnote{For example, we use the term \textit{Classification} for per-image class assignment while it implies a per-pixel class assignment in RS.}.
In addition, we distinguish between \textit{Segmentation} models that process a pixel at a time ('Segmentation - Pixel') and those that process the patch as a whole ('Segmentation - Patch'), as the throughput requirements of the latter are significantly higher.
Compared to the first column, the rest of the taxonomy uses RS terminology, as it focuses on the goals of the experiments, i.e., the RS-relevant information they generate.}
\CL{In the following group of columns, dedicated to the \textit{RS Problem}, experiments are hierarchically grouped by data modality (\textit{'Mod.'}), i.e., the type of RS data used as input, \textit{'Application'}, and \textit{'Dataset'}.}
As a result, experiments from the same study may not be juxtaposed.
\textit{'Application'} refers to the downstream task tackled by the study; applications are closely tied with the \textit{'Dataset'}\footnote{Datasets carrying the \textit{(cust.)} tag are unpublished datasets, meaning that the authors built their dataset but did not name or publish it.} and are detailed later in Fig.~\ref{fig:applications_grouped}.
The \textit{'Article'} group provides \CL{the experiment's} context: \textit{'Ref.'} lists the source article and \textit{'Year'} indicates its publication date.

The rest of the table is dedicated to the ML component of the experiment. 
\textit{'ML Model'} characterizes the model’s architecture through three fields: \textit{'Original Name'}, the exact appellation used in the study; \textit{Core}, a high-level architectural descriptor (e.g., CNN, Shallow NN such as a Multi-Layer Perceptron (MLP), or Traditional ML like an SVM), further detailed in Fig.~\ref{fig:models_per_year}; and, for Deep Neural Networks (DNNs), their \textit{'Backbone'}\CL{, i.e., the core of the network responsible for extracting features}.
Next, the \textit{'FPGA'} column indicates the family of the FPGA device used for implementation, more details in Fig.~\ref{fig:fpga_distribution}.
The last group of columns \textit{'Performance'} \CL{gathers}\CLstrike{reports} a few metrics of the experiment \CL{as reported in the original study}: \textit{'Score'} corresponds to the accuracy in \textbf{[\%]} of the model on its dataset and \textit{'MF\textbf{[MB]}'} is the Memory Footprint of the model in MegaBytes.
Finally, \textit{'C\textbf{[OP]'}} stands for the computational complexity of the model in numbers of operations. 
These performance metrics are the most closely related to the ML model; further metrics are reported during the second literature taxonomy in Table~\ref{table:fpga_optim}.

\subsection{Formulating RS applications as ML problems \RQinColor{applications}}\label{section:RQ1_RS_applications} 
From \CLstrike{urban planning to}disaster assessment \CL{to} target monitoring, RS images support a wide and growing range of EO applications \cite{zhaoOverviewApplicationsEarth2022}.

\begin{figure}
    \centering
    \includegraphics[trim={0 3.2cm 0 0.2cm},clip,width=0.7\textwidth] {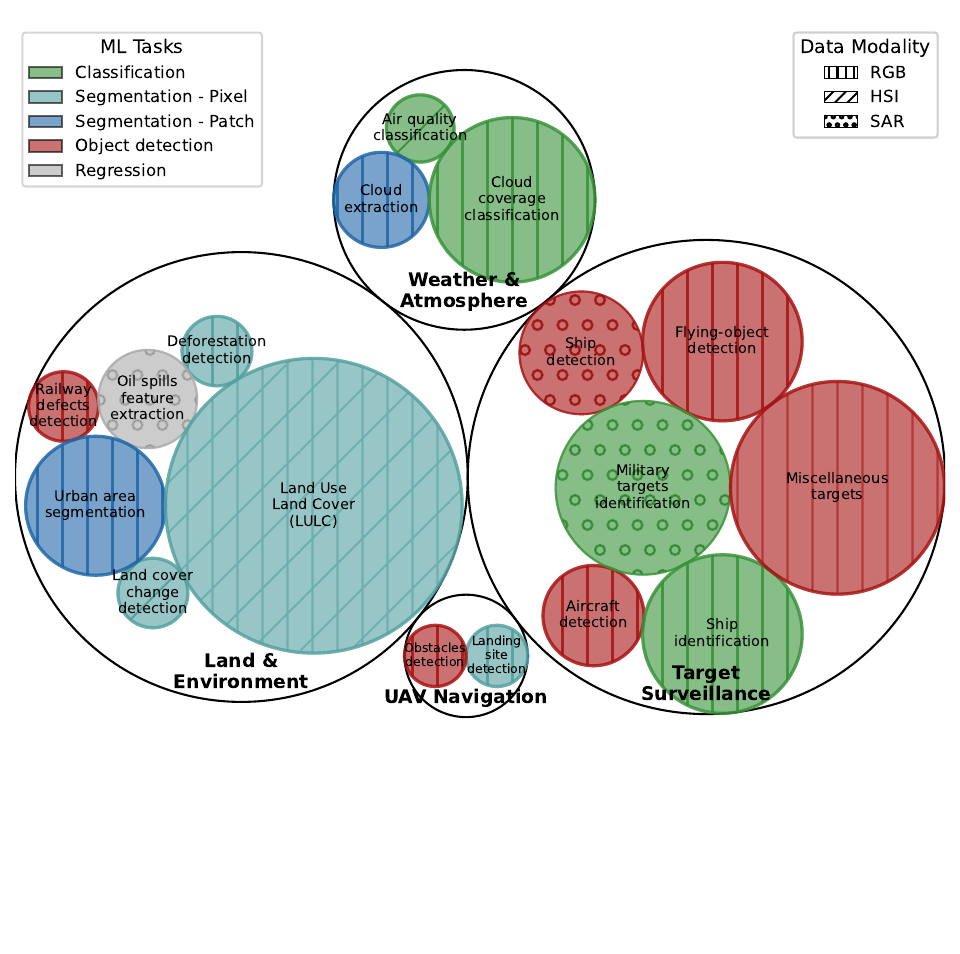}
    \Description[Bubble chart grouping RS applications by thematics]{
    A bubble chart showing applications of Remote Sensing data, colored by corresponding formulation as an ML task.
    The main categories are Land \& Environment, Weather \& Atmosphere, Target Surveillance, and UAV Navigation, represented by large transparent circles containing applications denoted by smaller circular patches.
    Land \& Environment includes several segmentation tasks for urban Areas or Land Use Land Cover, but also railway defects detection or oil spills feature extraction.
    Weather \& Atmosphere applications classify Cloud coverage or air quality, or extract clouds from images.
    Target Surveillance addresses aircraft detection, military target identification, ship identification or detection, flying-object detection, and miscellaneous targets mainly through object detection and classification techniques.
    UAV Navigation is the smallest circle; it involves segmenting safe UAV landing sites and detecting UAV obstacles.
    The chart also features a legend indicating which remote sensing data modalities (RGB, HSI, SAR) are used in each application.}
    \caption{Addressed RS applications grouped by thematic and colored as their corresponding ML task. The circles' size represents the number of experiments, while the hatching pattern represents the prevailing data modality.}
    \label{fig:applications_grouped}
\end{figure}

\paragraph{Overview of RS Applications}
Fig.~\ref{fig:applications_grouped} separates the surveyed RS applications into four distinct thematics.
\textit{Target Surveillance}\footnote{\textit{Target detection} is a more frequent appellation for RS applications. However, we will prefer the term "surveillance" to avoid any confusion with the ML task "object detection," sometimes also called "target detection."} gathers all applications focused on recognizing a specific family of objects.
The output of such a task can be the location---bounding box---of the targets, in such a case we denote it detection \CLstrike{(det.)}, or the binary presence of the target in the observed image, denoted identification \CLstrike{(id.)}.
In this thematic, "Miscellaneous targets" refers to heterogeneous object families, such as ships and cars, but also tennis courts or roundabouts in the DOTAv1.0 \cite{xiaDOTALargeScaleDataset2018} or DIOR \cite{liObjectDetectionOptical2020} datasets.
\CL{"Flying-object detection" and "Aircraft detection" differentiate objects in the air (planes, drones, birds, etc.) and standing aircraft on the ground.}
\CL{\textit{Land \& Environment} applications monitor the type of terrain present in the image, either on a general level, such as Land Use Land Cover (LULC), or for specific phenomena, such as oil spills.
In particular, "Land cover change detection" performs semantic segmentation of the land in an unsupervised fashion to identify pixels with abnormal values.}
\CLstrike{\textit{Landscape analysis} applications investigate the type of terrain present in the image.
In this survey, "Urban areas" refers to segmenting city patches, e.g., with the POTSDAM dataset \cite{2DSemanticLabeling}. 
"Landcover/Land use" is synonym to the classification of the type of terrain, such as desert, stadium, sea ice, etc., for example, in NWPU-RESISC45 \cite{chengRemoteSensingImage2017}.}
\CL{ In \textit{Weather \& Atmosphere}, we distinguish experiments that extract cloud masks from images and experiments that classify complete scenes based on cloud coverage thresholds.}
\CLstrike{Finally, \textit{Environment monitoring} includes landscape analysis applications specifically focused on environmental phenomena, such as cloud or deforestation detection.} 
\CL{Finally, \textit{UAV Navigation} applications rapidly use the acquired images to support the UAV's itinerary.}
While the grouping of Fig~\ref{fig:applications_grouped} offers a clear structure for the surveyed studies, it does not capture the full range of RS applications.
In particular, the records miss prevalent applications such as natural disaster monitoring or studies of the climate crisis. 

\paragraph{Data Modalities in RS}
The hash pattern in Fig.~\ref{fig:applications_grouped} represents the modality of the images used for each application.
In RS, the modality refers to the type of sensor used to acquire the data.
The most common modality, as in this survey, is optical data\footnote{\CL{Optical data is commonly named Multispectral in RS. However, as $\sim\!95\%$ of the surveyed studies only use the RGB channels of the electromagnetic spectrum, we use the term optical or RGB for accuracy.}}, with the three traditional Red-Green-Blue (RGB) channels. 
In RS, channels are commonly referred to as bands.
Indeed, sensors capture data around specific wavelengths, each wavelength's spectrum resulting in a different band.
Hyperspectral Imaging (HSI) emerges as the most extreme case, sometimes with over 200 bands, providing a very fine-grained spectral resolution that approximates a contiguous spectrum \cite{lopezPromiseReconfigurableComputing2013}.
Lastly, Synthetic Aperture Radar (SAR) is an active sensor, in opposition to passive sensors that collect light from the observed scene.
SAR emits waves that scatter upon hitting the observed scene, then captures the reflected components that bounce back towards the sensor.
Such a system enables data collection during nighttime and through clouds, making SAR a versatile modality.

\paragraph{Distinctions Between ML Tasks}
In Fig.~\ref{fig:applications_grouped}, each application circle is colored depending on the ML task it is primarily formulated as.
An ML task is the type of prediction made by the model. 
\CL{The most traditional tasks are \textit{Classification} and \textit{Regression}.
Also referred to as \textit{Scene Classification} and \textit{Parameter Retrieval} in RS terminology, the former assigns a category to each input image, while the latter predicts a numerical value from its features.}
\CLstrike{The most traditional tasks are \textit{Classification} and \textit{Regression}. \textit{Classification} assigns a category to each input image, while \textit{Regression} predicts a numerical value from its features.}
Beyond these, a Clustering task groups data instances into clusters based on shared characteristics.
When applied to images, Clustering yields a segmentation map, leading us to prefer the term \textit{Segmentation}. 
\CL{Semantic segmentation maps can be generated by processing the entire input image as a whole or by processing pixels one by one repetitively across the image.}
\CLstrike{\textit{Pixel Classification} also produces a segmentation map by individually processing and categorizing each pixel.
However, \textit{Pixel Classification} typically uses pixel-level features (e.g., spectral values, texture) as input, whereas \textit{Segmentation} also considers contextual information and spatial relationships between pixels.}
Finally, \textit{Object Detection} not only assigns categories but also provides bounding boxes indicating the location of objects, enabling the identification and categorization of multiple objects within a single image.
While some applications are closely tied to a specific ML task, others are more versatile and can be formulated in various ways.
Fig.~\ref{fig:applications_grouped} includes one example:
\CL{"LULC", was formulated once as \textit{Segmentation - Patch}, six times as \textit{Classification}, and 12 times as a \textit{Segmentation - Pixel} task.}\CLstrike{"Landcover/Land use," formulated five times as a \textit{Pixel Classification} task and four times as a \textit{Classification} task, and "Cloud Coverage Classification," expressed once as a \textit{Segmentation} task.}
Ultimately, how RS applications are formulated as ML tasks reflects the decisions and priorities of \CL{each study}\CLstrike{the individual studies}.

\paragraph{Converting RS Problems into ML Solutions} 
\textit{Target Surveillance} is a major focus in the literature.
Though uncommon in satellite imagery due to the typical low resolution of the images, this trend might be explained by the proportion of airborne data.
\CL{Indeed, Unmanned Aerial Vehicles (UAVs) offer higher-resolution imagery, which supports higher detection accuracy.}\CLstrike{Indeed, Unmanned Aerial Vehicles (UAVs) enable better resolution and, subsequently, higher detection accuracy.}
To mitigate low accuracies, many experiments express \textit{Target Surveillance} applications as scene-wise \textit{Classification}, a solution relying on prudent patchification of the original full-size RS image\footnote{
RS images are too large for most ML models to process directly. The process of patchification divides them into smaller, more manageable patches.}. 
Furthermore, we notice a significant focus on \CL{estimating cloud coverage.}\CLstrike{"Cloud coverage detection."}
Such an application is motivated by the importance of clouds in optical images.
Indeed, throughout the year, clouds cover over $66\%$ of Earth’s surface \cite{vitoloRealTimeOnboardSatellite2024}, significantly reducing the information obtainable from RGB images\footnote{Conversely, SAR wavelengths are specifically chosen to be outside the absorption spectrum of clouds, allowing radar signals to pass through without significant attenuation.}. 
Regarding ML tasks, most surveyed studies frame their applications as supervised \textit{Classification}.  
\CL{In contrast, \textit{Regression} appears only twice, making it the most under-represented task, although retrieving parameters from a scene is a critical problem in RS.}\CLstrike{In contrast, \textit{Regression} appears only once, making it the most under-represented task.}
Similarly, most datasets used for \CL{Segmentation - Pixel}\CLstrike{\textit{Pixel classification}} are labeled and result in supervised problems, except for the \CL{LULC}\CLstrike{"Landcover/Land use"} unsupervised classification of \citet{gyaneshwarRealtimeSCSUP2022}.
The \CL{semantic segmentation}\CLstrike{\textit{Segmentation}} of urban areas follows the same trend, though \citet{ratnakumarHighSpeedRoller2021} use unsupervised clustering. 
Finally, all \textit{Object detection} applications use multi-class datasets and single-shot models to detect multiple instances in a unique pass.


\begin{keymessage}{applications}
{\CL{Fully updated.} Target surveillance is the prominent focus of the surveyed literature ($44\%$). Frequently simplified as scene-wise classification, $31\%$ of the applications address defense concerns.
Landscape and environment monitoring dominates the rest of the research ($40\%$).
Regarding data modalities, RGB data is dominant ($65\%$), with HSI remaining niche and SAR primarily used for water environments.} 
\end{keymessage}




\subsection{Diversity and Trends in ML Models \RQinColor{models}}\label{section:RQ2_ML_models} 
\begin{figure}
    \centering
    \includegraphics[trim={0 0.45cm 0 0.3cm},clip,width=0.75\textwidth]{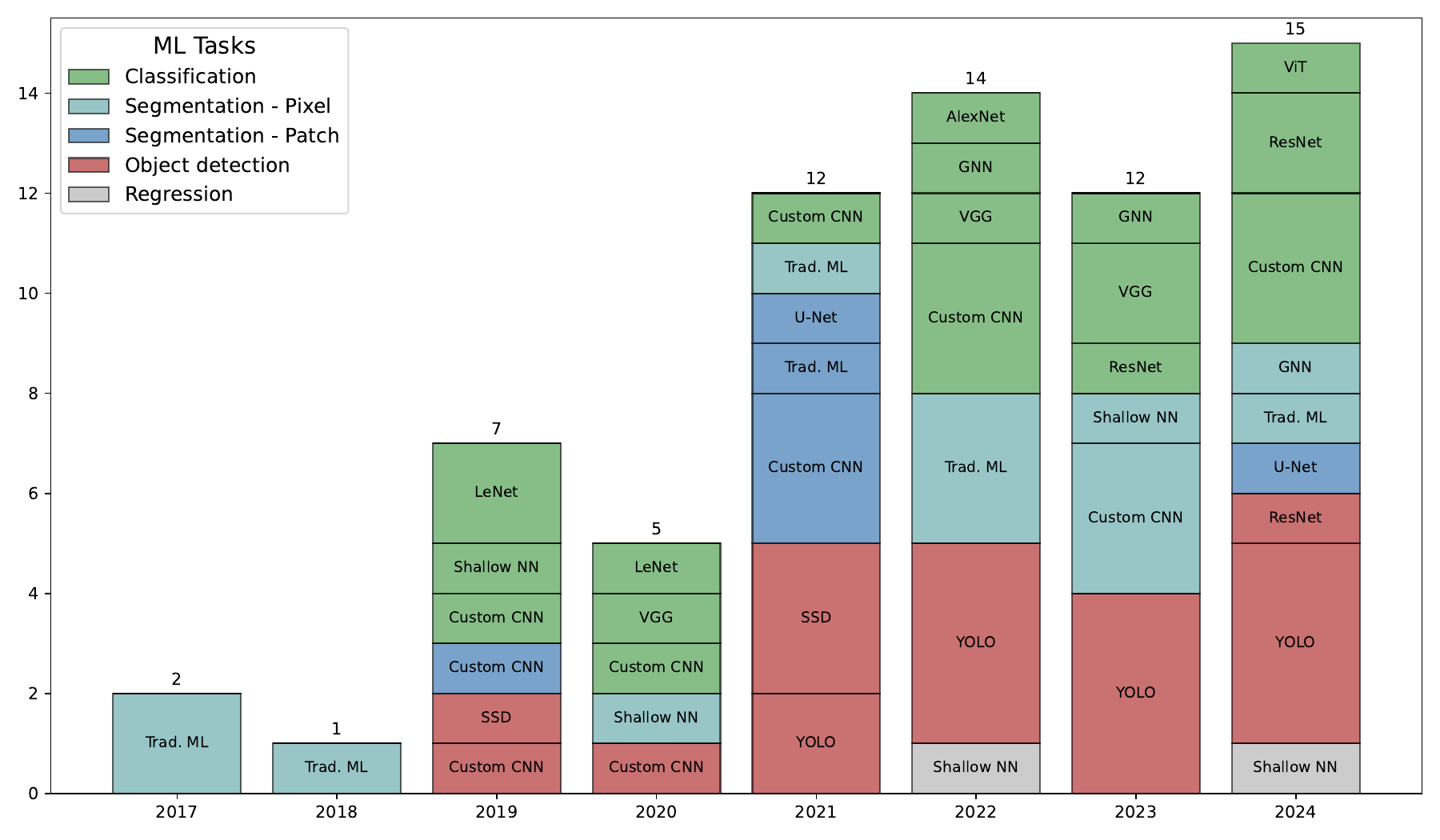}
    \Description[Task and ML model type distribution over the years]{A stacked bar chart showing the distribution of different ML tasks and model types over time (x-axis from 2017 to 2024).
    The tasks are Classification, Pixel Classification, Segmentation, Object Detection, and Regression, each represented by a different color.
    The y-axis represents the number of experiments.
    Each bar represents a year, while each sub-bar corresponds to the models grouped per "core".}
    \caption{ML model core architectures grouped per year and colored by their corresponding tasks}
    \label{fig:models_per_year}
\end{figure}

With \RQ{applications}, we analyzed the wide spectrum of RS applications and discussed how each experiment translates into an ML task. 
In \RQ{models}, we study the diversity of models used to interpret this information. 
Fig.~\ref{fig:models_per_year} shows the yearly distribution of models from the surveyed studies, categorized by ML task and labeled by each model’s "core", i.e., its key architectural characteristics.
Although this survey includes studies from 2014 onward, the earliest publication appears in 2017, with significant exploration beginning in 2019. 
From this point onward, the number of experiments and the diversity of architectures steadily increased, reflecting growing interest in the domain.

\paragraph{Evolution of ML Approaches to RS Problems} 
\CL{Surprisingly, among semantic segmentation tasks, more models operate per-pixel rather than processing the input patch at once,}\CLstrike{Surprisingly, more applications are expressed as \textit{Pixel classification} rather than \textit{Segmentation},} despite the latter being the modern approach for pixel-level analysis \cite{asgaritaghanakiDeepSemanticSegmentation2021}.
Indeed, \CLstrike{\textit{Segmentation}}\CL{segmentation} networks, like U-Net \cite{ronnebergerUNetConvolutionalNetworks2015}, have access to spatial information and generally outperform \CL{methods limiting spatial neighborhood information.}\CLstrike{\textit{Pixel classification} methods.}
However, resource constraints and computation overload also drive the model choice. 
Unlike \CL{\textit{Segmentation - Patch} deep} architectures, \CL{\textit{Segmentation - Pixel} experiments} often employ shallow networks or traditional ML, which are not only less demanding computationally but also easier to implement and parallelize. 
This combination of efficiency and simplicity likely explains their prevalence in this FPGA-focused survey, despite the broader RS trend favoring \textit{Segmentation - Patch}.
The remaining problems are formulated as \textit{Classification} and \textit{Object detection} tasks. Together, these two categories represent \CLstrike{over}two-thirds of the experiments. 
Such a trend can be explained by the versatility of formulating problems as \textit{Classification} tasks, as well as the importance of target monitoring in RS.

\paragraph{Examining Model Core Architectures} 
Model cores are chosen to highlight the fundamental differences between architectures.
In particular, traditional ML methods, such as SVMs \cite{cortesSupportvectorNetworks1995} or Decision Trees \cite{quinlanInductionDecisionTrees1986}, are isolated from Neural Networks (NNs), just as Shallow Networks are distinguished from Deep Neural Networks (DNNs) \cite{lecunDeepLearning2015}.
We split CNNs---the vast majority of the models analyzed in this survey---between renowned architectures, such as LeNet \cite{lecunGradientbasedLearningApplied1998}, AlexNet \cite{krizhevskyImageNetClassificationDeep2012}, VGG \cite{simonyanVeryDeepConvolutional2015}, U-Net \cite{ronnebergerUNetConvolutionalNetworks2015}, or ResNet \cite{heDeepResidualLearning2016}, and custom CNN architectures.
Notably, "Custom CNN" includes both newly designed CNNs and named architectures used only once in this survey, such as SqueezeNet \cite{iandolaSqueezeNetAlexNetlevelAccuracy2016} or ENet \cite{paszkeENetDeepNeural2016}.
Finally, we classify YOLO (v2, v3, v4) \cite{redmonYOLO9000BetterFaster2017, redmonYOLOv3IncrementalImprovement2018, bochkovskiyYOLOv4OptimalSpeed2020} and SSD \cite{liuSSDSingleShot2016} architectures separately, even though renowned CNN architectures are frequently used as the backbones of detection networks.
Indeed, the detection layers added to the feature extraction network (Fully Connected (FC) layers for YOLO and convolution layers for SSD) introduce fundamental differences to the architectures, such as the multiple heads of YOLO networks.
\CLstrike{Such characteristics reinforce the specialization of these models for object detection tasks.}

\paragraph{A CNN-Focused Landscape} 
Although shallow networks, such as MLPs, and traditional methods, like Decision Trees and SVMs, are featured in some studies \cite{gyaneshwarRealtimeSCSUP2022, martinsRealtimeSVMbasedHardware2024}---often due to their compatibility with resource-limited hardware---, the vast majority of contributions leverage DL algorithms.
Within DL, convolutional architectures are strongly preferred\CLstrike{: $76\%$ of all models are CNN-based}. 
Models like AlexNet, ResNet, and VGG are occasionally used, often for benchmarking purposes, but custom-designed CNNs are the most prevalent ($27\%$ of all models), reflecting designers' needs for medium-sized architectures typically with $6$ to $10$ layers. 
All detection networks are single-shot architectures like YOLO and SSD.
Two-stage detection alternatives like R-CNN \cite{girshickRichFeatureHierarchies2014a} are not used in the analyzed experiments.
Indeed, even if YOLO and SSD architectures are typically larger than R-CNN, the absence of a second pass through the networks highly reduces the computational complexity and processing time, making single-shot networks efficient solutions for resource-limited platforms like FPGAs \cite{zhaoHardwareAccelerationSatellite2023a}. 

\ifx\undefined\finalmanuscript

\paragraph{Selecting the Right Model} 
\CLstrike{Model selection is influenced by several factors, including the intrinsic suitability of architectures for specific tasks (e.g., YOLO and SSD for object detection) and hardware constraints.
For example, while SVMs can fit on most FPGA boards, larger models like ResNet50, with $\sim25M$ parameters and high memory and computational demands, are unsuitable for smaller devices.
This drives the prioritization of lightweight architectures, such as MobileNets \cite{howardMobileNetsEfficientConvolutional2017, sandlerMobileNetV2InvertedResiduals2018, howardSearchingMobileNetV32019}, and explains the large number of custom CNN designs \cite{liEfficientObjectDetection2019a, zhaoHardwareAccelerationSatellite2023a}.
Lastly, the software used to integrate and deploy the method on the FPGA platform also drives the model choice.
Indeed, if the designers are unfamiliar with low-level FPGA design, only the architectures already supported by higher-level toolchains are available.
We further explore this aspect later, through \RQ{FPGA-framework} in Section~\ref{section:fpga_taxonomy}.}

\fi
\begin{keymessage}{models}
{With a steadily growing number of publications per year, formulating RS applications as \textit{Classification} problems is prevailing (\CL{$35\%$}).
Of all the models, \CL{$74\%$} are convolution-based, of which \CL{$26\%$} are custom CNNs with an average of $9$ layers.
All detection models are single-shot networks and \CL{$89\%$} of the traditional ML methods are used in \CL{\textit{Segmentation - Pixel}} tasks.} 
\end{keymessage}

\subsection{The Role of the Computing Platform \RQinColor{FPGA-motivations}}\label{section:RQ3_FPGA_boards} 
As explored in the previous sections, deploying AI-supported RS applications in resource-constrained environments requires formulating the application as a suitable ML task and selecting an appropriate model.
These decisions are deeply intertwined with the selection of the hardware computing platform.
\CLstrike{These decisions are not only deeply intertwined with the selection of the hardware platform but also driven by project requirements.}
\CLstrike{For example, space deployment introduces harsh environmental conditions, including Single-Event Effects (SEEs) in onboard electronics caused by radiation exposure \cite{langeMachineLearningSpace2024}, tight power budgets from $20$W to $95$W for SmallSats payloads \cite{kothariFinalFrontierDeep2020}, and extreme temperature swings from $-150^\circ$C to $+150^\circ$C \cite{gordoSystemSpaceMaterials2020}---all of which place stringent demands on such edge computing hardware.}
\CLstrike{In light of these challenges, this section explores FPGAs' specific attributes and suitability for deploying AI on the edge compared to other hardware platform alternatives.}

\paragraph{Hardware Constraints Across RS Platforms}

\CL{
RS platforms are vehicles that carry remote sensors; they can operate at various altitudes as spaceborne, airborne, or ground-based systems.
Each platform operates under distinct hardware constraints.
Size, Weight, Power, and Cost (SWaP-C) define the fundamental trade-offs between computational capability and mission feasibility.
Additionally, payloads may require radiation hardening or thermal management.
For example, spaceborne systems face the most stringent constraints.
They are exposed to harsh environmental conditions, including Single-Event Effects (SEEs)\footnote{SEEs manifest as Single-Event Upsets (SEUs)---soft errors altering memory---or as Single-Event Latch-ups (SELs)---potentially destructive hard errors requiring a power reset.} in onboard electronics caused by radiation exposure \cite{langeMachineLearningSpace2024}, extreme temperature swings from $-150^\circ$C to $+150^\circ$C \cite{gordoSystemSpaceMaterials2020}, and tight power budgets from $20$W to $95$W for SmallSats payloads \cite{kothariFinalFrontierDeep2020}.
These extreme conditions make spaceborne systems the most constrained environment for FPGA-based ML deployment and, accordingly, the primary focus of the surveyed studies.
Airborne platforms, targeted by $\sim\!25\%$ of the articles, present moderate constraints that differ significantly between sub-categories \cite{wattsUnmannedAircraftSystems2012}.
Micro and nano UAVs operate at low altitudes and have short endurance, typically requiring light payloads and possibly high power if taking off vertically.
On the other hand, high- and medium-altitude aircraft can typically accommodate substantial payloads and possess longer endurance.
Finally, ground-based systems operate with minimal SWaP-C constraints and are absent from this review. 
Given these platform-specific limitations, this section examines how FPGAs address such challenges and how they compare with conventional computing hardware for edge AI deployment.
}

\begin{figure}
    \centering
    \includegraphics[width=0.8\textwidth]{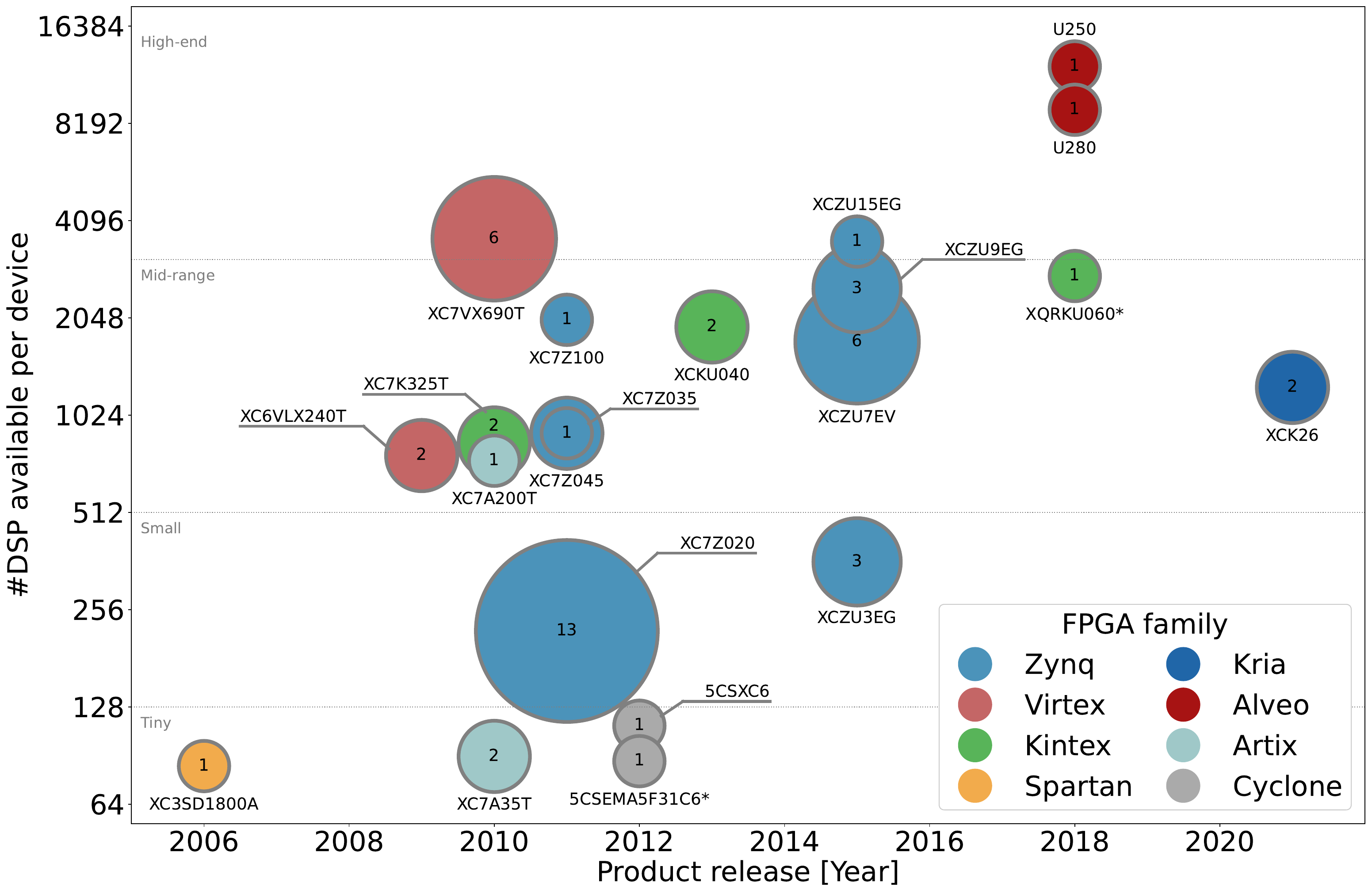}
    \Description[Bubble chart displaying FPGA boards used in article per year]{
    A bubble chart showing the usage of different FPGA-based products over time.
    The x-axis represents the release year of the FPGA board, and the y-axis represents the number of DSPs available per device on a logarithmic scale.
    Each bubble represents a specific FPGA product, its color indicating the FPGA family (Zynq, Kria, Virtex, Alveo, Kintex, Artix, Spartan, or Cyclone).
    The size of each bubble is proportional to the number of times the board was used in the surveyed articles. 
    The detailed FPGA part name is indicated under each bubble, and the legend shows the color-coding for each FPGA family.
    The bubbles follow a diagonal trend, indicating that newer boards tend to have more DSPs available.
    Horizontal dotted lines differentiate between four board sizes: Tiny (under 128 DSPs), Small (up to 512 DSPs), Mid-range (up to 3100 DSPs), and High-End (above).
    These 4 categories are used for discussion in \RQ{synergies}.}
    \caption{Distribution of FPGAs used across articles.
    Unlike other figures/tables, it shows FPGA products used per article ($48$) instead of per experiment ($68$).
    \textbf{*} indicates additional boards used in the studies, absent from the experiments reported in Table~\ref{table:fpga_optim}.
    The horizontal dotted lines represent arbitrary distinctions between size categories used in RQ7.}
    \label{fig:fpga_distribution}
\end{figure}

\paragraph{FPGA Usage in RS}\label{section:fpga_overview} 
FPGAs are reconfigurable integrated circuits, that offer design flexibility at the cost of programming complexity.
Often available off-the-shelf, FPGAs can operate independently on the edge or connect to other computational platforms via network cables or PCIe.
Fig.~\ref{fig:fpga_distribution} presents the different FPGAs used in each study. 
The devices are spread along the x-axis depending on their year of release, while the y-axis represents the number of DSP slices present on the board---relevant for arithmetic computations, like ML.
Except for one experiment using the Cyclone V FPGA from Intel, we observe that all considered studies deploy hardware from AMD (formerly Xilinx).
The most common family \CLstrike{with $60\%$}are Zynq FPGAs\CL{, used by $57\%$ of the studies}, which package FPGA programmable logic with CPUs on a single die to support networking and general-purpose tasks. 
Such products will be referred to as System-on-Chip (SoC) FPGAs. 
In contrast, datacenter FPGAs, such as the Alveo Series, are often offered as PCIe cards that can be inserted into a server.
\ds{Furthermore, to address the need for ionizing-radiation tolerance to ensure mission reliability, FPGA vendors offer radiation-hardened and radiation-tolerant variants. 
AMD's KCU105 kit, for example, is a commercial equivalent of the space-grade Kintex Ultrascale XQRKU060 used in the VIDEO project \cite{nerisFPGABasedImplementationCNN2022a}}.
In addition to FPGAs, designers can select from several families of Processing Units (PUs) when deploying AI models on edge devices\dsstrike{each offering unique advantages}.
In the following subsections, we look at the alternatives: CPUs, GPUs, and ASICs (custom, TPUs, and VPUs).
We analyze the authors' motivations to select FPGA over alternatives and mention, when available, the comparisons made by the studies.
\ds{In the end, we summarize the unique advantages of each family of PUs in Table~\ref{table:HW_platforms_comparison}}.

\paragraph{FPGAs vs. CPUs}
Central Processing Units (CPUs) are ubiquitous general-purpose processing units, widely available as Commercial Off-The-Shelf (COTS) components.
While well-suited for complex data transformations and system management, CPUs are limited in parallel processing capabilities, often falling behind specialized hardware in raw computational power.
Evaluating FPGA-based designs against CPUs is a common practice in the surveyed studies.
These comparisons often involve desktop and server processors, including older Intel i7 models (up to 8th gen) \cite{martinsRealtimeSVMbasedHardware2024,hashimotoShipClassificationSAR2019a,liEdgeRealtimeObject2023a,yangLightweightDetectionMethod2023}, newer i7/i9 models (12th and 13th gen) \cite{huangEdgeTrustworthyAI2024,zhangAcceleratingGNNbasedSAR2023}, AMD Ryzen 3990x \cite{zhangAccurateLowlatencyEfficient2022a}, and Broadwell Xeon server CPUs \cite{niAlgorithmHardwareCoOptimization2023,zhangEfficientFPGABasedImplementation2020,wuDesignImplementationRemote2021}.
While these desktop and server CPUs offer high computational performance, they are not designed for energy-efficient applications\dsstrike{, making direct comparisons with FPGA implementations less straightforward}. 
In contrast,\dsstrike{some studies include} mobile CPUs, such as the AMD 4800H \cite{zhangAccurateLowlatencyEfficient2022a,zhaoHardwareAccelerationSatellite2023a},\dsstrike{which} provide a more power-conscious alternative.
Similarly, \citet{huangEdgeTrustworthyAI2024} compare their ETAUS system against an embedded ARM Cortex A530 CPU and achieve significant performance improvements---$\times 25$ GOP/s/W and $\times 37$ FPS.
%
\ds{Superior power efficiency, compared to both CPUs and GPUs, is the most frequently cited advantage of FPGAs \cite{yangLightweightDetectionMethod2023, gyaneshwarRealtimeSCSUP2022, yanAutomaticDeploymentConvolutional2022a, heConfigurable2D3D2023a, liEdgeRealtimeObject2023a, zhangFPGAImplementationCNNbased2021a, chenHardwareImplementationConvolutional2020, bahlLowpowerNeuralNetworks2019a, liNovelCNNBasedAP2DNet2020, matos-carvalhoStaticDynamicAlgorithms2019}.
For instance, \citet{yangLightweightDetectionMethod2023} reports $29\times$ higher efficiency compared to an Intel Core i7-8700 K, and \citet{groverReductionPowerConsumption2012} demonstrate different design techniques to optimize power consumption.}
\dsstrike{
Independently of direct comparison to CPUs, one of the most frequently cited advantages of FPGAs is their superior energy efficiency \cite{yangLightweightDetectionMethod2023, gyaneshwarRealtimeSCSUP2022, yanAutomaticDeploymentConvolutional2022a, heConfigurable2D3D2023a, liEdgeRealtimeObject2023a, zhangFPGAImplementationCNNbased2021a, chenHardwareImplementationConvolutional2020, bahlLowpowerNeuralNetworks2019a, liNovelCNNBasedAP2DNet2020, matos-carvalhoStaticDynamicAlgorithms2019}.
For instance, \citet{yangLightweightDetectionMethod2023} report a power efficiency ratio $29$ times superior to an Intel Core i7-8700 K. 
Indeed, FPGAs' power consumption can be optimized using various design techniques, as detailed by \citet{groverReductionPowerConsumption2012}.}
\dsstrike{Furthermore, FPGAs are often favored for their radiation tolerance, with vendors like AMD offering radiation-hardened and radiation-tolerant Virtex and Kintex products.}
Although some CPUs and microcontrollers, such as ARM Cortex-R or LEON families, are radiant-tolerant, FPGAs' inherent reconfigurability facilitates the implementation of mitigation techniques, like Triple Modular Redundancy (TMR), where critical logic is triplicated and majority voting used to mask errors. 
On the downside, \dsstrike{radiation-hardened}\ds{rad-hard} FPGAs are significantly more expensive than rad-hard CPUs and microcontrollers.
Their SRAM-based memory is also vulnerable to SEUs, requiring mitigation techniques \cite{sabogalMethodologyEvaluatingAnalyzing2021a,siegleMitigationRadiationEffects2015}.
\dsstrike{While investigating SRAM sensibility,}\citet{siegleMitigationRadiationEffects2015} highlight that, although CPUs and microcontrollers remain essential onboard, especially for the OBC, modern payload processing applications increasingly rely on FPGAs.
Over time, new solutions to deploy ML algorithms on embedded devices have been developed, shifting the focus from CPUs to domain-specific accelerators \cite{raoofyBenchmarkingMachineLearning, shawahnaFPGABasedAcceleratorsDeep2019}.

\paragraph{FPGAs vs. GPUs}
Graphics Processing Units (GPUs) possess many small cores, enabling high parallelism and performance for tasks involving massive amounts of data and large matrix computations.
They are widely used for training NNs.
Consequently, many studies compare FPGA-based designs against high-performance NVIDIA GPUs.
These include Maxwell \cite{wuDesignImplementationRemote2021,hashimotoShipClassificationSAR2019a} and Pascal architectures \cite{niAlgorithmHardwareCoOptimization2023, zhangEfficientFPGABasedImplementation2020, zhangExtremelyPipelinedFPGAbased2023a, yanAutomaticDeploymentConvolutional2022a, zhangFPGAImplementationCNNbased2021a, liEfficientObjectDetection2019a}, which introduce native half-precision support.
Later architectures, such as Volta \cite{liEdgeRealtimeObject2023a, myojinDetectingUncertainBNN2020} and Turing \cite{nguyenFPGASoCImplementationYOLOv42024, zhaoHardwareAccelerationSatellite2023a, yangLightweightDetectionMethod2023, yangAlgorithmHardwareCodesign2022}, contain Tensor Cores, processing units dedicated for Deep Learning operations.
Ampere GPUs \cite{yuImprovedLightweightDeep2024,zhangAccurateLowlatencyEfficient2022a,zhangAcceleratingGNNbasedSAR2023} introduce native support for low precision integer operations, thus heavily benefiting from quantization.
The latest Blackwell GPUs \cite{NVIDIABlackwellArchitecture} support precision as low as 4-bit floating-point, decreasing FPGAs' benefits gained from using custom datatypes. 
\dsstrike{GPUs' considerable processing capabilities come at the cost of significant memory and power consumption---often exceeding $300$W \cite{collangePowerConsumptionGPUs2009}---making them impractical for most edge deployments.}\ds{FPGAs can exhibit better power efficiency \cite{weiFPGABasedHybridTypeImplementation2019} compared to desktop GPUs.
For example, \citet{liEdgeRealtimeObject2023a} report $134\%$ of the FPS of a Tesla V100 while consuming $10 \times$ less power.
In contrast, \citet{liEfficientObjectDetection2019a} find that an NVIDIA GTX 1070 Ti outperforms a Zynq-7000 in GOP/s/W efficiency by a factor of three showing, that power efficiency is highly implementation-dependent.
Regardless of efficiency, GPUs' considerable processing capabilities come at the cost of significant memory and power consumption---often exceeding $300$W \cite{collangePowerConsumptionGPUs2009}---making them impractical for most edge deployments.}
\dsstrike{To address these constraints, edge GPUs offer improved efficiency for AI inference.
Computing boards, like the NVIDIA Jetson Nano \cite{NVIDIAJetsonNano}, integrate CPUs with energy-efficient GPUs microarchitectures.}\ds{To address these constraints, computing boards, like the NVIDIA Jetson Nano \cite{NVIDIAJetsonNano}, combine CPUs with energy-efficient edge GPUs. 
Even though edge GPUs usually deliver higher peak performance \cite{liEfficientObjectDetection2019a}, FPGAs can still achieve better power efficiency \cite{weiFPGABasedHybridTypeImplementation2019}.
For example}\dsstrike{Yet}, \citet{huangEdgeTrustworthyAI2024} report that their FPGA-based system ETAUS achieves $2.7 \times$ GOP/s/W compared to Jetson Nano.
Similarly, the AP2D-Net presented by \citet{liNovelCNNBasedAP2DNet2020} achieves $2.8 \times$ FPS/W compared to the best model of the DAC System Design Contest 2018 \cite{xuDACSDCLowPower2021} deployed on a Jetson TX2.\dsstrike{Notably, \citet{liEfficientObjectDetection2019a} compare their FPGA implementations against the theoretical peak performance of edge GPUs like the Jetson AGX Xavier.
They highlight that a significant gap in achievable performance still exists, even after extensive design optimization.}\dsstrike{While GPUs deliver high computational throughput, FPGAs typically exhibit better power efficiency in edge scenarios \cite{weiFPGABasedHybridTypeImplementation2019}.
For instance, \citet{liEdgeRealtimeObject2023a} report achieving $134\%$ of the FPS of a Tesla V100 while consuming $10 \times$ less power, whereas \citet{liEfficientObjectDetection2019a} find that an NVIDIA GTX 1070 Ti outperforms a Zynq-7000 in GOP/s/W efficiency by a factor of three.
These variations highlight that power efficiency is highly implementation-dependent.} The complex architecture and high transistor density of GPUs pose significant challenges for circuit-level radiation hardening.
While the NVIDIA Tegra K1\dsstrike{ (TK1) has demonstrated the ability to withstand} \ds{has withstood} radiation levels present in short-duration LEO missions \cite{badiaReliabilityEvaluationLU2022}, no edge GPU, to the authors' knowledge, has been explicitly designed for space-grade reliability.\dsstrike{This is in contrast to the previously mentioned radiation-tolerant FPGAs. The need for ionizing-radiation tolerance to ensure mission reliability is more readily addressed by the accessibility of radiation-tolerant FPGAs \cite{upadhyayDesignImplementationCNNbased2024}. 
AMD's KCU105 kit, for example, is a commercial equivalent of the space-grade Kintex Ultrascale XQRKU060 used in the VIDEO project---covered by \citet{nerisFPGABasedImplementationCNN2022a}.}
In summary, while GPUs---especially edge GPUs---offer strong computational performance and power efficiency trade-offs for ML tasks, FPGAs remain the preferred choice in power- and space-constrained environments due to their energy efficiency and radiation tolerance.

\paragraph{FPGAs vs. ASICs (TPUs/VPUs)}
Application-Specific Integrated Circuits (ASICs) are fully custom-designed chips optimized for specific applications, potentially offering unmatched performance and energy efficiency.
Yet, their design is a complex, lengthy, and expensive process---no study has compared an FPGA-based design against a suitable custom ASIC.
Many studies list reconfigurability as a key advantage of FPGAs over ASICs \cite{ratnakumarHighSpeedRoller2021, yangLightweightDetectionMethod2023, torresCombinedWeightlessNeural2020, zhangFPGAImplementationCNNbased2021a, liNovelCNNBasedAP2DNet2020, yahiaouiParallelizationFuzzyARTMAP2017a}; in defense and space for example, re-programmability enhances mission versatility by enabling the "deploy and program" paradigm\footnote{Allowing post-launch updates and in-orbit reconfiguration.} \cite{heConfigurable2D3D2023a, liEdgeRealtimeObject2023a}.
Consequently, FPGAs are increasingly replacing traditional ASICs due to their faster time-to-market, cost-efficiency in small-batch applications, and competitive performance \cite{rapuanoFPGAbasedHardwareAccelerator2021a}.
Despite these advantages, ASICs remain the optimal choice for ultra-low-power applications, as the generic nature of FPGAs cannot match the efficiency of fully customized hardware \cite{amaraFPGAVsASIC2006}. 
%
%
Tensor Processing Units (TPUs) are commercially available ASICs optimized for Deep Learning, primarily used in data centers \cite{jouppiTPUV4Optically2023}.
While the Google Coral TPUs \cite{CoralTPUDevBoardDatasheet} target edge devices, no study in this survey directly compared FPGA designs to a TPU.
\citet{liEfficientObjectDetection2019a} only compare their Zynq 7000 design against Google cloud TPU’s theoretical peak performance.
Vision Processing Units (VPUs), like Intel's Movidius Myriad series, focus on low-power CNN inference.
However, compared to the Myriad 2 VPU \cite{yangAlgorithmHardwareCodesign2022}, FPGAs hold a key advantage in radiation tolerance.
In a direct comparison, \citet{rapuanoFPGAbasedHardwareAccelerator2021a} build on the CloudScout initiative \cite{giuffridaCloudScoutDeepNeural2020} and conclude that the Myriad 2 VPU is suitable only for short Low Earth Orbit (LEO) missions due to radiation susceptibility---unlike the space-grade Kintex Ultrascale XQRKU060 they also evaluate and is SEL-immune.
Performance-wise, their design shows a $2.4 \times$ speed-up compared to the VPU, but with $1.8 \times$ higher power consumption and a longer development time.

\begin{table}
\caption{\ds{Summary of different computing platforms used in the different studies.}}
\label{table:HW_platforms_comparison}
\small
\begin{tabular}{lccccccc}
    \toprule
    Chip &  Dev. Time & Price/Unit & Rad-Hard (Avail.) & Updatable & Pow. Eff. & Latency  & Peak Perf. (\textit{fp32})\\
    \midrule	
    \textbf{CPU }& Short    & Low           & Yes   & Yes   & Low       & Medium    & Medium \\
    \textbf{GPU }& Medium   & Medium/High   & No    & Yes   & Medium    & High      & Very high\\
    \textbf{Custom ASIC}& Very long& Very low      & Yes   & No    & Very High & Very Low  & Custom\\
    \textbf{FPGA}& Long     & High          & Yes   & Yes   & High      & Low       & Medium \\
    \bottomrule
\end{tabular}
\end{table}

\paragraph{A Large Catalog of Computing Platforms}
\ds{Through the motivations of the surveyed studies, and in line with the conclusions of \citet{wangDeepNeuralNetwork2020} and \citet{lentarisHighPerformanceEmbeddedComputing2018}, a generic guideline for hardware platform selection, considering development effort and performance targets, is reported in Table~\ref{table:HW_platforms_comparison}.
GPUs, TPUs, and VPUs offer quicker paths to achieving initial performance, especially for computationally complex DNNs \cite{bayerReachingEdgeEdge2024}.
FPGAs, however, promise superior performance and efficiency if more development effort and hardware expertise are invested in custom accelerator design and optimization.
For applications demanding ultimate performance and willing to bear the highest development costs and longest design cycles, custom ASICs represent the pinnacle, albeit with reduced updatability \cite{amaraFPGAVsASIC2006}.
In addition, both ASICs and FPGAs can directly connect their compute units to input and output ports, enabling lower latencies compared to GPUs and CPUs.
In the following sections, we examine the technical aspects of the surveyed studies, exploring the methodologies and techniques employed to implement and optimize ML models on FPGAs for RS applications.}

\dsstrike{Through the motivations of the surveyed studies, and in line with the conclusions of \citet{wangDeepNeuralNetwork2020} and \citet{lentarisHighPerformanceEmbeddedComputing2018}, a generic guideline for hardware platform selection, considering development effort and performance targets, can be formulated.
While many hardware platforms are available to accelerate AI inference on the edge, only a few COTS systems support heavier models like DNNs.
Examples include NVIDIA's Jetson Nano GPU \cite{NVIDIAJetsonNano}, Google's Coral TPU \cite{CoralTPUDevBoardDatasheet}, and Intel's Myriad 2 VPU \cite{IntelMovidiusMyriad}, all presenting solid alternatives for terrestrial edge computing.
However, their susceptibility to radiation limits their use in space applications, where FPGAs offer a more resilient alternative \cite{rapuanoFPGAbasedHardwareAccelerator2021a}.
While floating-point operations on FPGAs may exhibit comparable---or worse, due to lower clock frequency---performance to other embedded platforms, the true advantage of FPGAs emerges when leveraging fixed-point, typically 8-bit, or even binary (1-bit) operations \cite{wangDeepNeuralNetwork2020}.
This capability, often not fully supported or efficiently implemented on other hardware platforms, allows FPGAs to achieve significant performance and efficiency gains.}

\dsstrike{
GPUs, TPUs, and VPUs offer quicker paths to achieving initial performance, especially for computationally complex DNNs \cite{bayerReachingEdgeEdge2024}.
FPGAs, however, promise superior performance and efficiency if more development effort and hardware expertise are invested in custom accelerator design and optimization.
For applications demanding ultimate performance and willing to bear the highest development costs and longest design cycles, custom ASICs represent the pinnacle, albeit with reduced flexibility \cite{amaraFPGAVsASIC2006}.
In the following sections, we look at the technical aspects of the surveyed studies, exploring the methodologies and techniques employed to implement and optimize ML models on FPGAs for RS applications.
}

\begin{keymessage}{FPGA-motivations}
{AMD FPGAs overwhelmingly dominate the surveyed hardware, comprising nearly all devices (\CL{$96\%$}). The preferred FPGAs are SoC boards, particularly the Zynq family ($57\%$).
Most boards are equipped to run DNNs (\CL{$60\%$} with $>500$ DSPs), and almost all devices have accelerated at least one DL model in an experiment.
Reconfigurability, power efficiency, and radiation concerns are the primary motivations for choosing FPGAs over other DNN-friendly COTS accelerators.}
\end{keymessage}

\section{From AI Models to FPGA Deployment: Design Strategies and Challenges}\label{section:design}
\subsection{Optimizing AI Models for Efficient Inference \RQinColor{optimizations}}\label{section:RQ4_optimizations} 
Before deploying a model on an FPGA, it typically requires adaptation and compression.
These optimization steps are closely linked to hardware design decisions---discussed \CL{later} in \RQ{FPGA-framework} and \RQ{FPGA-designs}.
In \RQ{optimizations}, we introduce key network compression techniques and discuss their impact and requirements.
Since traditional ML methods are already lightweight, we primarily discuss optimizations for computationally intensive models, like NNs.

\paragraph{Architectural Optimizations via Lightweight Backbones}
Most NNs include a feature extractor, or backbone, responsible for refining and compressing the input's essential information. 
Extracting rich, high-quality features often relies on large and deep backbones, which come with high computational costs.
To lighten these costs, designers can exploit lightweight backbones.
For instance, \citet{yangLightweightDetectionMethod2023} replace YOLOv3’s DarkNet53 backbone with GhostNet \cite{hanGhostNetMoreFeatures2020}, reducing the number of learnable parameters by a factor of $3$.  
Similarly, \citet{liEdgeRealtimeObject2023a} swap YOLOv4’s CSPDarknet53 backbone for MobileNeXt \cite{zhouRethinkingBottleneckStructure2020}, lowering the parameter count from $60.08$M to $36.44$M, with only a $1.65\%$ mAP drop---compensated via other customizations. 
As discussed in \RQ{models}, the existence of many architectures make selection a challenging task.
\citet{sabogalMethodologyEvaluatingAnalyzing2021a} compare four DL segmentation models, showing that ENet \cite{paszkeENetDeepNeural2016} outperforms U-Net \cite{ronnebergerUNetConvolutionalNetworks2015} while being $23.81\times$ more computationally efficient and requiring $20.56\times$ less memory.  
Similarly, \citet{yangAlgorithmHardwareCodesign2022} compare three popular lightweight backbones: MobileNetv1 \cite{howardMobileNetsEfficientConvolutional2017}, MobileNetv2 \cite{sandlerMobileNetV2InvertedResiduals2018}, and SqueezeNet \cite{iandolaSqueezeNetAlexNetlevelAccuracy2016}, concluding that MobileNetv1 achieves the best trade-off for their SAR ship detection application.  
MobileNetv1 also introduces a width multiplier for layer-wise scaling, leveraged by \citet{suhAlgorithmHardwareCoOptimizationEnergyEfficient2021} to explore different deployment scenarios in terms of accuracy and FPGA resource use.  
Beyond swapping architectures, \citet{kimOnOrbitAICloud2024} propose a multi-phase cloud coverage detection pipeline where images first undergo a uniformity check and classification via a small CNN (TriCloudnet) before full segmentation with U-Net.
Such workflow avoids the computation of cloud-clear or overcast images, reducing processing time and power consumption by $48.7\%$.
When sacrificing some feature representation capabilities is manageable, lightweight and multi-phase approaches highlight that questioning a model’s depth and width can result in significant computational savings. 


\paragraph{Reducing the Complexity of Convolutional Layers} 
Given the ubiquity of convolution operations in Computer Vision, lightweight backbones commonly focus on efficiently computing them.
For instance, MobileNets architectures \cite{howardMobileNetsEfficientConvolutional2017, sandlerMobileNetV2InvertedResiduals2018, howardSearchingMobileNetV32019} replace standard convolution layers by depth-wise separable convolutions. 
This form of factorized convolutions separate the standard convolution operation in two stages: filtering, with single 2D filters for each input channel, and combination, by linearly combining the output of the filtering with a 1x1 convolution. 
When considering an input tensor $h_i \cdot w_i \cdot d_i$, a kernel $k \cdot k \cdot d_i \cdot d_j$, and an output tensor $h_i \cdot w_i \cdot d_j$, a standard convolution costs $h_i \cdot w_i \cdot d_i \cdot d_j \cdot k \cdot k$, while \CL{a} depth-wise separable convolution costs $h_i \cdot w_i \cdot d_i (k^2 \cdot d_j)$, a computational reduction of a factor $\approx k^2$.
For example, \citet{howardMobileNetsEfficientConvolutional2017} use depth-wise separable convolution to reduce computation by $9\times$ for a loss of $1\%$ on ImageNet. 
As an alternative path to convolution approximation, All Adder Networks \cite{chenAdderNetWeReally2020} replace the multiplications in convolution kernels by additions.
While it does not reduce computational complexity, additions are generally less expensive than multiplications.
In particular, the All Adder NN of \citet{zhangExtremelyPipelinedFPGAbased2023a} achieves \SI{3}{} TOP/s, the highest computational throughput of the survey. 


\paragraph{Pruning Techniques for Compact Networks} 

Pruning reduces computational complexity and memory costs by removing non-essential weights or entire channels from an NN. 
While the usual overparameterization of NNs aids training, it results in many redundant weights \cite{szeEfficientProcessingDeep2017}.
After training, pruning nullifies the least significant weights (in contribution to the output) to reduce this redundancy, but often requires fine-tuning to recover lost accuracy.
The resulting trade-off between parameter reduction and task accuracy can be adjusted with a pruning factor, which determines the percentage of weights to be nullified.
To further enhance pruning efficiency, sparsity is often enforced during training via techniques like L1-regularization \cite{zhangAccurateLowlatencyEfficient2022a, zhaoHardwareAccelerationSatellite2023a}.
For example, \citet{zhangAccurateLowlatencyEfficient2022a} train their GNN with lasso regression---adding an L1-penalty term to the loss---to encourage weight shrinkage and facilitate pruning.
Channel pruning removes entire feature channels, enabling more efficient hardware execution.
\citet{yangLightweightDetectionMethod2023} prune Ghost-YOLO channels, reducing the number of parameters from \SI{23.52}{M} to \SI{6.71}{M} at a cost of \SI{2.93}{\%} mAP\footnote{High accuracy costs are usually mitigated with other optimizations, in this case via Knowledge Distillation (KD) \cite{hintonDistillingKnowledgeNeural2015}, discussed further in Section~\ref{section:discussion}.}. 
Similarly, \citet{nguyenFPGASoCImplementationYOLOv42024} prune \SI{60}{\%} of filter weights, incurring only a \SI{0.85}{\%} mAP loss.
Beyond weights and channels, certain families of models also have the opportunity to prune input data. 
For example, \citet{zhangAccurateLowlatencyEfficient2022a} reduce the computational complexity of their GNN by \SI{92.8}{\%} by pruning input graph vertices, causing only a \SI{0.17}{\%} OA drop.
To leverage the efficiency improvements reached through pruning, \citet{yangAlgorithmHardwareCodesign2022} perform hardware-guided progressive pruning.
The method integrates an initial coarse pruning\CLstrike{, mixed precision Quantization-Aware Training (QAT)}, a resource-per-layer cost estimation, and a final resource-aware fine pruning.
Pruning is frequently paired with quantization for further efficiency gains \cite{hanDeepCompressionCompressing2016a, zhaoHardwareAccelerationSatellite2023a}.


\paragraph{Custom Datatypes Through Quantization} 
One of FPGA’s key computational advantages over alternative hardware platforms, like GPUs, lies in custom datatype operations.
Quantization reduces numerical precision by converting full-precision 32-bit floating-point (\textit{fp32}) parameters into lower-precision formats such as 16-bit fixed-point (\textit{i16}), 8-bit integer (\textit{i8}), or even binary (\textit{b}).
Low-bandwidth datatypes enable highly efficient arithmetic operations, particularly on FPGA logic, where binary multiplications can be performed using simple XOR operations. 
While quantization reduces computational cost and memory footprint, it inevitably perturbs model parameters,\footnote{Several factors influence the effectiveness of a quantization scheme, such as its uniformity, symmetry, and granularity. 
For a more detailed discussion on quantization techniques, we recommend the comprehensive survey of \citet{gholamiSurveyQuantizationMethods2021}.} shifting them from their convergence point and causing accuracy degradation compared to full-precision implementations \cite{gholamiSurveyQuantizationMethods2021}.
For instance, \citet{sabogalMethodologyEvaluatingAnalyzing2021a} report an average accuracy drop of $1\%$ across four segmentation models.
Similarly, in object detection, \citet{zhaoHardwareAccelerationSatellite2023a} observe a $0.42\%$ drop in mAP, while \citet{niAlgorithmHardwareCoOptimization2023} report a \CL{minimal reduction in classification accuracy}\CLstrike{classification accuracy reduction} of $0.03\%$ for VGG16 and $0.06\%$ for ResNet-34. 
Although the quantization-induced loss rarely exceeds $1\%$, it can be unacceptable in scenarios requiring high model reliability.
To mitigate accuracy degradation, designers may opt to re-train the network, effectively choosing one of two standard approaches: Post-Training Quantization (PTQ) or Quantization-Aware Training (QAT).  
PTQ, sometimes called direct quantization, is straightforward to implement as it applies quantization directly after training.
However, PTQ's simplicity can cause significant accuracy degradation, especially in models with a wide parameter distribution \cite{jacobQuantizationTrainingNeural2018}.  
For instance, \citet{weiFPGABasedHybridTypeImplementation2019} observe over $50\%$ accuracy loss across all tested bit widths when applying PTQ to a LeNet-5 for MSTAR classification.
In contrast, QAT simulates low-precision arithmetic already during training, resulting in better robustness to quantization noise \cite{pitonakCloudSatNet1FPGABasedHardwareAccelerated2022}, albeit at a higher computational cost \cite{yuImprovedLightweightDeep2024}. 
Beyond PTQ and QAT, mixed-precision quantization assigns different bit widths to different sections of the model.  
This method preserves accuracy by keeping sensitive layers at higher precision while applying aggressive quantization to less critical layers.
For example, \citet{yangAlgorithmHardwareCodesign2022} identify inefficient layers by estimating layer-wise latency and resource consumption. 
\citet{rapuanoFPGAbasedHardwareAccelerator2021a} take this idea further by manually assigning custom bit widths to each input, filter, and output, achieving a $48\%$ memory footprint reduction for CloudScout.
Alternatively, some approaches like \citet{zhangEfficientFPGABasedImplementation2020} mitigate accuracy loss by selectively retaining \textit{fp32} precision for activations while quantizing only the weights.  
Similarly, \citet{weiFPGABasedHybridTypeImplementation2019} introduce a symmetric QAT method that quantizes weights but preserves normalization and activation in \textit{fp32} format.  
The resulting computation overhead is motivated by the need to preserve the model's accuracy.
Quantization is indispensable to achieve FPGA designs that are competitive with higher-frequency computing platforms. 
While PTQ is often sufficient to achieve 8-bit quantization with minimal accuracy degradation \cite{nagelWhitePaperNeural2021}, QAT is preferable when maintaining accuracy is critical.  
Mixed-precision quantization further optimizes efficiency by balancing accuracy and computational cost.  

\paragraph{Combining Optimization Techniques} 
All the methods discussed above effectively optimize NNs, but their combined use often yields the best results.
\CL{Notably, combining lightweight backbones and fixed-point quantization is a recurring and effective strategy in the surveyed studies \cite{yangLightweightDetectionMethod2023, yangAlgorithmHardwareCodesign2022, yuImprovedLightweightDeep2024, liEdgeRealtimeObject2023a, liEfficientObjectDetection2019a, nerisFPGABasedImplementationCNN2022a, nguyenFPGASoCImplementationYOLOv42024, zhaoHardwareAccelerationSatellite2023a, kimOnOrbitAICloud2024}.
Additionally, lightweight backbone replacement or the direct use of efficient architectures, such as ENet \cite{paszkeENetDeepNeural2016} or \textit{“tiny”} variants like YOLOv4-tiny \cite{yuImprovedLightweightDeep2024, nguyenFPGASoCImplementationYOLOv42024}, appear in $\sim\!20\%$ of the surveyed experiments, offering an effective trade-off between accuracy and complexity.
When the accuracy drop must be compensated, designers typically enhance the baseline architecture or training setup.
For example, \citet{nguyenFPGASoCImplementationYOLOv42024} add a third YOLOv4 head to detect targets at multiple spatial scales and optimize training with data augmentation to recover accuracy losses.
Several surveyed studies \cite{zhaoHardwareAccelerationSatellite2023a, nguyenFPGASoCImplementationYOLOv42024, hanDeepCompressionCompressing2016a} combine backbone replacement, pruning, quantization, and a recovery method in an incremental optimization strategy.
Such gradual development approaches allow designers to evaluate the contribution of each step in an ablative manner, progressively refining both accuracy and resource efficiency before final FPGA deployment.}
\CLstrike{
For example, \citet{zhaoHardwareAccelerationSatellite2023a} analyze in an ablative fashion the parameters and accuracy reduction of each subsequent technique applied.
The resulting figure allows to clearly track which method provides the highest parameter reduction---backbone replacement in their case---and which method helps recover lost accuracy---here, data augmentation.
Similarly, \citet{yangLightweightDetectionMethod2023} follow a structured design flow to progressively refine YOLOv4.
They first replace its backbone with GhostNet \cite{hanGhostNetMoreFeatures2020} (leveraging depth-wise separable convolutions), then apply channel pruning to drastically reduce parameters.
To restore accuracy, they incorporate Knowledge Distillation (KD) \cite{hintonDistillingKnowledgeNeural2015}, before finally quantizing the model.
This multi-step process results in a model with just $\approx 10\%$ of the original parameters and $5\%$ of the memory footprint.
A similar approach is presented by \citet{hanDeepCompressionCompressing2016a}, who combine pruning, quantization, and Huffman coding (weight clustering) to achieve a $35\times$ storage reduction on AlexNet and $49\times$ on VGG16, all without accuracy loss.} 
Ultimately, the effectiveness of these optimization techniques depends not only on compression gains but also on their alignment with FPGA accelerator design, discussed in the next section.

\begin{keymessage}{optimizations}
{To reduce computational load, \CL{24}\CLstrike{26}\% of DNNs leverage lightweight backbones or efficient architectures, primarily MobileNets (\CL{46}\CLstrike{43}\%). 
Out of all DL experiments, \CL{89}\CLstrike{91}\% use quantization, mostly to \textit{int$8$} format, and \CL{26}\CLstrike{26}\% leverage pruning to compress their model. 
Fine-granularity mixed-precision quantization presents significant opportunities for further network compression.}
\end{keymessage}




\subsection{Finding the best framework to efficiently use FPGAs \RQinColor{FPGA-framework}}\label{section:fpga_taxonomy} 

Implementing NNs on CPUs and GPUs typically involves using stable and mature software frameworks and drivers.
In contrast, automatic deployment frameworks for FPGAs have not seen the same level of adoption---most implementations depend on manual frameworks and varying design patterns.
To support future work in selecting suitable implementation frameworks and design patterns, we provide a dedicated second taxonomy in Table~\ref{table:fpga_optim}.
It presents key metrics relevant to the respective FPGA implementation, \ds{as reported in the original study,} organized into three column groups.
First, \textit{Implementation Choices} includes the implementation framework \textit{'Impl.'}, the model family \textit{'Fam.'}, the design pattern \textit{'P'} (\textit{S} for Specific, \textit{F} for Flexible), the \textit{'FPGA'} product, and the \textit{'Model Name'}.
Second, \textit{Design Metrics} covers the precision post quantization \textit{'Prec.'}, the model complexity \textit{'C\textbf{[OP]}'}, and the memory footprint \textit{'\textbf{[MB]}'} (\textbf{\textdagger}: model parameters are located on-chip, and not loaded from external memory).
Additionally, we report the DSP \textit{'D\textbf{[\%]}'} and BRAM \textit{'B\textbf{[\%]}'} resource utilization.
Finally, \textit{Performance Metrics} attempts to compare different designs with each other.
In particular, we report the operating frequency in \textit{'\textbf{[MHz]}'}, computational throughput\dsstrike{Footnote: See Section~\ref{section:recommendations} for a discussion about the confusion between \textbf{[GOP]}, \textbf{[GOPs]}, and \textbf{[GOP/s]}.}
\textit{'T\textbf{[GOP/s]}'}, power consumption \textit{'P\textbf{[W]}'}, computational efficiency \textit{'T/P'}, Latency \textit{'L\textbf{[s]}'}, and temporal throughput in \textit{'\textbf{[FPS]}'}.

\begin{table}
\centering

\caption{FPGA Optimization Taxonomy Table}
\label{table:fpga_optim}
{\tiny
\begin{tabular}{cccccccclccccccccc}
 \multicolumn{7}{c}{\textbf{Implementation Choices}} & \multicolumn{5}{c}{\textbf{Design Metrics}} & \multicolumn{5}{c}{\textbf{Peformance Metrics}}  \\
 \cmidrule(lr){1-7}  \cmidrule(lr){8-12} \cmidrule (lr){13-18}

&\textbf{Impl.}&\textbf{Fam.} &\textbf{P} &\textbf{Ref.} &\textbf{FPGA} &\textbf{Model Name} &\textbf{Prec.} &\textbf{C[OP]}&\textbf{MB}  &\textbf{D[\%]} &\textbf{B[\%]} &\textbf{MHz} &\textbf{T[GOP/s]} & \textbf{P[W]} & \textbf{T/P} &\textbf{L[s]} &\textbf{FPS}\\
 \toprule
 \multirow{39}{*}{\rotatebox[origin=c]{90}{\textbf{Manual}}}
   &\multirow{27}{*}{HDL}
       &\multirow{19}{*}{CNN}
           &\multirow{6}{*}{S}
               &\multirow{1}{*}{\cite{ieracitanoExplainableEmbeddedNeural2024}}
                   &\multirow{1}{*}{7Z045}
                       &\multirow{1}{*}{HO-ShipNet}
                           &\multirow{1}{*}{i16}
                               &\multirow{1}{*}{-}
                                   &\multirow{1}{*}{-}
                                       &\multirow{1}{*}{44}
                                           &\multirow{1}{*}{44}
                                               &\multirow{1}{*}{-}
                                                   &\multirow{1}{*}{-}
                                                       &\multirow{1}{*}{1.90}
                                                           &\multirow{1}{*}{-}
                                                               &\multirow{1}{*}{1.01 ms}
                                                                   &\multirow{1}{*}{-}\\
\bottomrule 
\end{tabular}

\flushleft{
\textit{FPGA:} AMD/XILINX Fpga names are stripped of leading  to increase readability\\
\textit{Mem:} \textbf{\textdagger} Uses On-Chip Memory, \textbf{?} Memory Location unknown\\
\textit{Latency:} \textbf{*} Latency of a single Pixel\\
\textit{Model Name:} \textbf{RDBC} Roller Dung Beetle Clustering, \textbf{WNS} Weightless Neural System\\
}
}
\end{table}


\paragraph{Implementation Choices}
The designs are split between manual implementations---Hardware  Description Language (HDL) and High-Level Synthesis (HLS)---and automatic frameworks---FINN \cite{umurogluFINNFrameworkFast2017}, Vitis AI \cite{AMDVitisAI}, MATLAB, and VGT. 
HDL is the most commonly used language ($\approx 45\%$), offering precise control of the generated circuit. However, most designs do not exploit this level of control to use all available resources nor push the theoretical limit of the clock frequency.
High-Level Synthesis (HLS) tools offer a more abstract approach to FPGA programming as they enable the use of high-level languages to create larger designs with less lines of code.
Even with HLS, a solid grasp of FPGA architecture and hardware concepts remains necessary, especially when compared to GPU programming environments, like CUDA. 
In the reported designs, we detect no overarching trend in worse resource utilization or clock frequency of HLS designs compared to HDL designs.
The average resource utilization of implementations are $\approx$30\% DSPs and 50\% BRAM for HDL and 46\% DSPs and 43\% for HLS with a high variance between individual designs.
Most manual designs focus on CNN-based networks ($\approx$ \ds{70}\dsstrike{71}\%), a well-studied network architecture commonly implemented using an array of Processing Elements (PEs) that pass intermediate results between each other.

To reduce the overhead of porting CNNs to FPGAs automatic frameworks have been developed.
The two most commonly used in this survey are AMDs Vitis AI and the FINN compiler.
The FINN compiler \cite{umurogluFINNFrameworkFast2017} automatically generates a model \textit{Specific} accelerator for CNNs.
Once implemented, the FINN design is solely useful for the specific network. It heavily depends on quantization to reduce the network complexity and memory footprint. \citet{myojinDetectingUncertainBNN2020} binarize a 4 layers CNN, and \citet{pitonakCloudSatNet1FPGABasedHardwareAccelerated2022} compare 2,3 and 4-bit quantization using FINN.
The Vitis AI compiler \cite{AMDVitisAI} supports the \textit{Flexible} design approach, using a Deep Learning Processor Unit (DPU) flashed to the FPGA able to run multiple different AI workloads.
The peak Operations Per Cycle (OPC) of the DPU can be customized and matched to the algorithm \cite{AMDVitisAI}.
\citet{sabogalMethodologyEvaluatingAnalyzing2021a} test multiple networks with different DPU configurations, demonstrating that some networks have utilization as low as $14.2\%$ (ESPNet) and up to $85.6\%$ (U-Net) using the B1024 (1024 OPC) architecture. 
Vitis AI can leverage multiple DPU cores to increase performance as demonstrated by \citet{nguyenFPGASoCImplementationYOLOv42024} and \citet{yuImprovedLightweightDeep2024}.
On the downside, Vitis AI requires PTQ or QAT, as it only supports the \textit{i8} datatype \cite{AMDVitisAI}. 
Other automated frameworks include MATLAB (2023a) \cite{gargAircraftDetectionSatellite2024} and the Xilinx System Generator (XSG) \cite{gyaneshwarRealtimeSCSUP2022}, but both works do not go into detail concerning their implementation.
\citet{bahlLowpowerNeuralNetworks2019a} use VGT, a VHDL compiler generating circuits for abstract CNNs similar to FINN; it has, however, no continuing support.
Frameworks that are not classified do not provide enough information on their implementation.



\paragraph{Design Pattern}  
We split each workload into two distinct design patterns. 
\textit{Flexible (F)} designs are bitstreams that \dsstrike{can} support multiple different kernels and network topologies without requiring reconfiguration. 
Supporting different networks and kernels usually requires weights to be loaded from off-chip memory.
\citet{zhangAccurateLowlatencyEfficient2022a} use a \textit{Flexible}  design with on-chip memory, pre-loading memory to the weight buffer, and amortizing the access over multiple classifications.
The support for different layers comes at a small cost in computational performance, but allows implementing larger models that would otherwise not fit on the FPGA using a \textit{Specific} approach.
The average footprint of the $20$ \textit{Flexible} designs is $25MB$ compared to \dsstrike{$15MB$}\ds{$14MB$} for the \dsstrike{$10$}\ds{$11$} \textit{Specific} designs.
Unlike \textit{Flexible} designs, \textit{Specific} designs only support a single network architecture.
Deploying a different model requires re-implementation of the bitstream or is not supported.
This allows the design to be hyper-specialized to the model, making full use of the customizability of the FPGA.
\textit{Specific} designs can use different datapaths for different pruning and quantization of individual layers.
Differentiating model families (\textit{'Fam.'}) is relevant as different ML models have vastly distinct requirements. 
For example, CNNs spend most of their time performing convolution, a task with high computational intensity \cite{rapuanoFPGAbasedHardwareAccelerator2021a}, whereas GNNs require a high memory bandwidth \cite{zhangAcceleratingGNNbasedSAR2023}.
The FPGA device also influences the performance, similar to CPUs, the number of transistors increases over the generations, enabling designers to use more resources such as DSPs, BRAM, and LUTs (see~\ref{section:fpga_overview}).

\paragraph{Design Metrics}
The \textit{'Prec.'} column reports the precision of operations post-quantization. The most popular quantization is \textit{i8}, with \dsstrike{$51\%$}\ds{$50\%$} of all implementations and \dsstrike{$44\%$}\ds{$43\%$} of manual ones using either \textit{i8} or a mixture of \textit{i8} with \textit{f32} or \textit{i32}.
Next, we report the model complexity \textit{'C\textbf{[OP]}'}, an essential metric as a higher complexity requires more computational resources to produce the same temporal throughput \textit{'FPS'}.
In addition, the relationship between model complexity and memory footprint can indicate whether a model is compute- or memory-bound.
Most of the designs do not reach the compute boundary nor provide an off-chip memory bandwidth analysis making a further discussion impossible.
The next metric is the model footprint and the location of parameters.
FPGAs, similarly to CPUs, contain restricted on-chip memory, e.g., $<5$MB for Zynq MPSoC chips, and a main off-chip memory ($512$MB to $32$GB). 
Using on-chip memory eliminates most memory bandwidth concerns as distributed BRAM provides very high bandwidth.
On-chip memory weights can either be stored in BRAM---pre-loaded before execution---or stored inside LUTs---disabling any change in parameters.
\textit{Specific} designs mostly use on-chip memory with only \dsstrike{$25\%$}\ds{$24\%$} using off-chip memory,  while $93\%$ of \textit{Flexible} designs use off-chip memory.
In particular, the FINN compiler stores weights on-chip, such as BRAM and LUTs, while Vitis AI loads weights from external memory.
Storing the weights on-chip leads to the design being mostly limited by BRAM \cite{wangAccelerationImplementationConvolutional2019,zhangExtremelyPipelinedFPGAbased2023a,fraczekEmbeddedVisionSystem2018}.
The average memory footprint of on-chip implementation is only \dsstrike{$0.83MB$}\ds{$0.72MB$} compared to $28MB$ for off-chip memory implementation.
The last two middle columns present the DSP \& BRAM utilization.
A high utilization can indicate a good selection of the model for the FPGA, while a low utilization can indicate bad scalability of the design.
This excludes network architectures that cannot use DSPs and BRAMs efficiently, such as BNNs---using LUTs for most operations---or GNNs---with low data locality and computational intensity.
\citet{yahiaouiParallelizationFuzzyARTMAP2017a} compare an implementation using only LUTs with a DSP- and BRAM-based implementation.
They show that the BRAM\_DSP implementation can double the number of pixel processed in parallel, while using the same amount of resources.

\paragraph{Performance Metrics}
Comparing different models on CPUs and GPUs is already difficult; for FPGAs, where the design can vary even more, finding a fair comparison is especially complex.
One common metric relevant for comparison is the computational throughput, marking the peak number of operations per second.
Nevertheless, operations with different precisions cannot be directly compared, as floating-point computations are more expensive than integer operations, and smaller bit widths yield higher \textbf{[GOP/s]}.
For example, the highest throughput is reported as $3$ TOP/s \cite{zhangExtremelyPipelinedFPGAbased2023a}.
However, the solution uses an All Adder NN, which approximates the functionality of a convolution filter by using only additions. 
Moreover, the intermediate feature maps resulting from the convolution layers are sent back to the CPU.
Thus, the Fully-Connected (FC) layers, corresponding to $92\%$ of the parameters, are executed on the CPU, drastically reducing the memory footprint. 
The remaining layers only represent $10M$ parameters, which can fit on the boards $6.6$MB BRAM due to the quantization to 4-bit. 
Using BRAM eliminates any memory bottlenecks and the main computation, a 4-bit adder, can be implemented using solely LUTs, resulting in the extreme \dsstrike{reported }throughput.
The second highest reported throughput is $452$ GOP/s \cite{liEfficientObjectDetection2019a}.
Compared to \citet{zhangExtremelyPipelinedFPGAbased2023a} the design uses 16-bit values loaded from memory, and the main operations are Multiply-Accumulate (MAC), counted as two operations.
The DSP block inside a 7-series Xilinx FPGA can perform one MAC operation per cycle and they manage to keep $1152$ DSPs ($60 \%$) busy achieving $1.97 \times 1152 \times 200MHz \approx 452 GOP/s$.
\citet{sabogalMethodologyEvaluatingAnalyzing2021a} and \citet{suhAlgorithmHardwareCoOptimizationEnergyEfficient2021} use double datarate DSPs to increase the performance---an optional optimization inside Vitis AI.
Doubling the clock frequency of DSP components, which can be clocked higher than the rest of the design, roughly doubles the available computational capability.
A majority of experiments ($\approx 72\%$) report the power consumed by the FPGA, a value pointless on its own as the total energy required to compute a sample is more important. 
To attempt a more informative comparison between different models and implementations, we calculate the computational efficiency \textit{'T/P'} in $[\frac{GOP/s}{W}]$.
\citet{zhangExtremelyPipelinedFPGAbased2023a} achieve the highest computational efficiency $368 \frac{GOP/s}{W}$ due to the high throughput detailed previously. 
On the other hand, \citet{heConfigurable2D3D2023a} have an extremely low $<2~\frac{GOP/s}{W}$ efficiency, even though they use a low-precision datatype (\textit{i8}).
Another key design metric is the frequency \textit{'MHz'}, which depends on the speed-grade and technology of the device, as well as the critical paths in the design.
A low frequency, compared to designs on the same FPGA, can hint at a poorly designed circuit with long critical paths.
\textit{'Latency'} and \textit{'FPS'} values are model- and implementation-specific, incomparable between different workloads and implementations\ds{, and only included}\dsstrike{Nevertheless, we include them in Table~\ref{table:fpga_optim}} to ensure completeness.
It is important to separate \textit{'Latency'} from \textit{'FPS'}, as, generally, multiple inputs can be processed inside the accelerator pipeline, increasing the FPS significantly compared to a non-pipelined implementation that can process one element at a time.

\begin{keymessage}{FPGA-framework}
{Extensive work in implementing CNNs on FPGAs has produced various automated frameworks supporting common layers.
Within the surveyed studies we could not find a trend showing better performance of manual designs in any of the metrics.
\textit{Specific} design patterns are commonly used ($\sim\! 75\%$) with limited on-chip memory, while \textit{Flexible} design patterns tend to support larger models ($+10$MB in average).
}
\end{keymessage}

\subsection{Optimizing Performance for FPGA implementations \RQinColor{FPGA-designs}}\label{section:RQ5_FPGA_design}

FPGAs have a reconfigurable architecture, which enables designers to implement energy-efficient dataflow architectures that re-use data, and are adapted to the compute operations of the algorithm.
However, FPGAs' reconfigurability comes at the cost of frequency---compared to CPUs and GPUs.
Therefore, the use of different kinds of parallelism is necessary to achieve similar performance.

Instruction Level Parallelism (ILP) exploits the execution of consecutive instructions in parallel, for example overlapping load and MAC operations \cite{suhAlgorithmHardwareCoOptimizationEnergyEfficient2021}.
This level of parallelism is exploited in almost every design and can be performed manually in HDL or automatically using HLS control directives for loop statements.
Consecutive iterations of a loop can overlap, increasing the throughput and amortizing the latency over multiple loop iterations.
In addition, ILP designs commonly exploit Data Level Parallelism (DLP), for example by executing multiple MAC operations in parallel to perform convolution \cite{yanAutomaticDeploymentConvolutional2022a}.
After optimizing individual tasks using ILP, designers can leverage Task Level Parallelism (TLP), for example, by overlapping different layers in a NN---a mechanic heavily exploited by FINN \cite{umurogluFINNFrameworkFast2017}.
The compiler connects individual layers with First-In First-Out (FIFO) streams and starts executing a layer as soon as its first inputs are produced by the previous layer.
Increasing the batch size increases TLP, enabling different inputs to be processed at the same time.
The disadvantage of larger batch sizes is a possible increase in latency, leading to most implementations using a batch size of $1$ during inference.

Increasing computation puts pressure on the memory bandwidth, and most designs use multiple PEs in a Systolic Array \cite{kungWhySystolicArchitectures1982} to reduce off-chip memory accesses.
\ds{Furthermore, transformer-based networks \cite{wickramasingheVTROptimizedVision2024}, relying on dense matrix multiplication with a lower computational intensity than convolutional layers, require more off-chip bandwidth.
Similarly, GNNs, exhibiting the lowest computational intensity, necessitate an even greater focus on creating an advanced network to reduce the number of accesses to off-chip memory as much as possible \cite{zhangAcceleratingGNNbasedSAR2023,zhangAccurateLowlatencyEfficient2022a}.}
\dsstrike{Furthermore}\ds{Given enough bandwidth}, multiple PEs enable simple scaling to use most of the FPGA resources; this approach can be found in many designs \cite{chellaswamyFPGAbasedRemoteTarget2024,heConfigurable2D3D2023a,liEfficientObjectDetection2019a,yangAlgorithmHardwareCodesign2022,pitonakCloudSatNet1FPGABasedHardwareAccelerated2022,rapuanoFPGAbasedHardwareAccelerator2021a,shibiOnboardTargetDetection2021a,suhAlgorithmHardwareCoOptimizationEnergyEfficient2021,wangAccelerationImplementationConvolutional2019,weiFPGABasedHybridTypeImplementation2019,wuDesignImplementationRemote2021,yahiaouiParallelizationFuzzyARTMAP2017a,yanAutomaticDeploymentConvolutional2022a,yangLightweightDetectionMethod2023,zhangAccurateLowlatencyEfficient2022a,zhangAcceleratingGNNbasedSAR2023,zhangExtremelyPipelinedFPGAbased2023a,zhangFPGAImplementationCNNbased2021a,zhangEfficientFPGABasedImplementation2020}. 
Using multiple levels of parallelism, most designs can be scaled to reach the computational limit for floating-point operations on FPGAs, which require multiple DSPs per operation.
To increase the \textbf{OP/s} available, designers can choose to reduce datatypes precision, hence reducing the number of resources required per operation.
Most designs use quantization to a fixed-point datatype to increase the number of available operations.
In addition to DSPs, which can perform basic arithmetic operations on integers, LUTs can perform any 6-bit to 1-bit mapping or a 5-bit to 2-bit operation.
Thus, a BNN or low-precision network can be mapped to LUTs directly, allowing for high peak \textbf{OP/s} \cite{zhangExtremelyPipelinedFPGAbased2023a}.
The trade-off between LUT and DSP utilization is automatically explored by \citet{yangAlgorithmHardwareCodesign2022}.
In addition to computational benefits, lower precision reduces off-chip memory requirements and can allow the network to fit entirely into BRAM.
Another option to increase available operations is to run DSPs at double the frequency, effectively doubling their throughput \cite{suhAlgorithmHardwareCoOptimizationEnergyEfficient2021,sabogalMethodologyEvaluatingAnalyzing2021a}.
\dsstrike{For networks with low computational intensity, the focus is placed on creating an advanced network to reduce the number of accesses to off-chip memory as much as possible \cite{zhangAcceleratingGNNbasedSAR2023,zhangAccurateLowlatencyEfficient2022a}.}

\CL{Finally, the large spatial scale of RS imagery also plays a decisive role in FPGA design, as it makes naive full-image processing impractical. 
Consequently, most implementations rely on patch-based strategies, which maintain throughput but introduce an additional pre-processing overhead---often negligible compared to models' complexity.
In parallel, differences in sensing modalities also shape algorithmic design choices.
For instance, HSI data, with hundreds of spectral bands, are frequently processed pixel-by-pixel \cite{gyaneshwarRealtimeSCSUP2022, martinsRealtimeSVMbasedHardware2024, chellaswamyFPGAbasedRemoteTarget2024, boyleHighlevelFPGADesign2023a, shibiOnboardTargetDetection2021a}, while others sample channels to decrease spectral redundancies and computational load \cite{heConfigurable2D3D2023a}.
In contrast, all studies handle SAR imagery as amplitude-only---effectively grayscale images---and therefore avoid any complex-valued operations or datatypes in the implementation.
}

\begin{keymessage}{FPGA-designs}{
Most manual designs use a combination of different forms of parallelism and quantization to increase the throughput of the design.
To reduce the off-chip memory bandwidth, designs can re-use data, commonly employing a Systolic Array of PEs.
Comparing the quality of different designs proves difficult, and most surveyed studies do not compare their CNN implementations with previous work.}
\end{keymessage}
\section{AI and FPGA Synergies: towards Next-Gen Earth Observation Missions}\label{section:onboard} 

\CL{In each experiment, designers must select an ML model, an FPGA device, and an accelerator architecture.
In this vast search space, finding the best AI and FPGA technology combinations constitutes a tedious process.
While the diversity of study setups makes any comparison challenging and generalization even more difficult, this section discusses promising associations between AI and FPGA technology.}

\subsection{Limits and suitabilities of AI-supported RS applications on FPGAs \RQinColor{synergies}}\label{section:RQ8_best_methods}

\CLstrike{While the setup of each study differs---making any comparison challenging and generalization even harder---we would like to highlight and discuss some of the best solutions surveyed\footnote{Dataset accuracies were compared to the benchmarks available at \url{https://paperswithcode.com/sota}}.}\ds{To demonstrate the feasibility of different ML models and RS applications for an increasing number of resources, we discuss trends among solutions based on the size of the FPGA.
We use the amount of DSPs available, displayed in Fig.~\ref{fig:fpga_distribution}, as a proxy for the size of the FPGA as other resources (LUTs, FFs, BRAM) scale in a similar fashion.
}

\begin{figure}
    \centering
    \includegraphics[width=\textwidth]{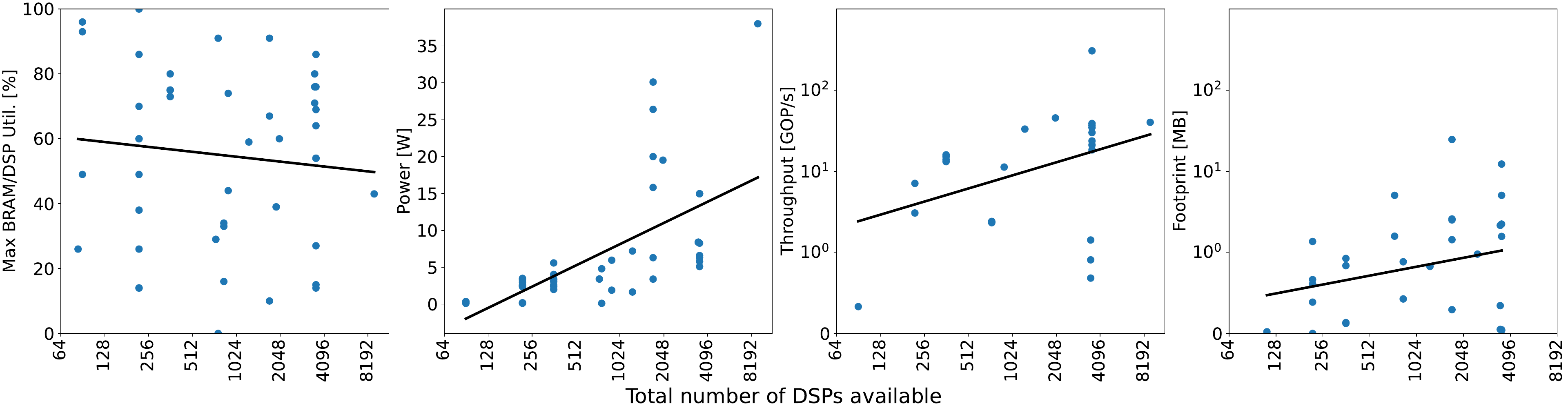}
    \Description[Four lineplots displaying performance metrics per FPGA sizes]{The figure shows 4 line plots displaying 4 performance metrics on y-axes against the total number of DSPs available.
    The first plot displays the maximum utilization of DSPs and BRAM in \%, and the line fitted using linear regression shows no correlation with the FPGA size.
    The second subplot presents the reported power, and the fit line grows steadily with larger FPGA devices.
    The third and fourth plots display the computational throughput and the memory board sizes, whose lines also grow with respect to the FPGA size.}
    \caption{Maximum utilization of DSP \textit{'D\textbf{[\%]}'} and BRAM \textit{'B\textbf{[\%]}'}, reported power \textit{'P\textbf{[W]}'}, computational throughput \textit{'\textbf{[GOP/s]}'}, and footprint \textit{'\textbf{MB}'} for different board sizes. We plot only reported metrics and fit the black line using linear regression.}
    \label{fig:FPGA_metrics_subplot}
\end{figure}

\paragraph{Tiny and Small Devices}
\CL{Among the four experiments using sub-150 DSP FPGAS, which we categorize as tiny, two designs deploy equally tiny models to perform pixel segmentation \cite{gyaneshwarRealtimeSCSUP2022} and image classification \cite{bahlLowpowerNeuralNetworks2019a}.
For its part, \citet{vitoloRealTimeOnboardSatellite2024} implement a larger network ($7403$ parameters) running at low frequency.
All these designs achieve marginal power consumption ($<0.5$W) and demonstrate the use case of tiny FPGA devices for traditional ML models and tiny CNNs for fast approximate classification.} 
\CL{The next size of FPGAs are small FPGA devices, often featured in educational evaluation boards.
Two FPGAs of this class are used: first, the XC7Z020 with 220 DSPs and DDR3, and second, the newer generation XCZU3EG with 360 DSPs and faster DDR4 memory. 
Both devices successfully deploy traditional ML or shallow NN designs, but also smaller CNN-based models are implemented to segment images \cite{sabogalMethodologyEvaluatingAnalyzing2021a, kimOnOrbitAICloud2024}, classify ships \cite{hashimotoShipClassificationSAR2019a} or clouds \cite{pitonakCloudSatNet1FPGABasedHardwareAccelerated2022},
or detect miscellaneous objects \cite{wangAccelerationImplementationConvolutional2019, yangLightweightDetectionMethod2023, suhAlgorithmHardwareCoOptimizationEnergyEfficient2021, liNovelCNNBasedAP2DNet2020, wangFastDetectionObstacle2024}.
Deploying CNNs on small FPGAs requires the use of reduced models \cite{yangLightweightDetectionMethod2023, pitonakCloudSatNet1FPGABasedHardwareAccelerated2022, wangFastDetectionObstacle2024} or sacrifices in FPS \cite{kimOnOrbitAICloud2024, yangLightweightDetectionMethod2023, wangAccelerationImplementationConvolutional2019, hashimotoShipClassificationSAR2019a}.
Interestingly, two studies \cite{sabogalMethodologyEvaluatingAnalyzing2021a, suhAlgorithmHardwareCoOptimizationEnergyEfficient2021} examine the trade-off between model size, throughput, and accuracy. 
\citet{suhAlgorithmHardwareCoOptimizationEnergyEfficient2021} achieve particularly impressive results using their custom 8-bit UniPOT quantization scheme and leveraging DSP double rate, and perform inference $< 2.6$W for all three SSD sizes.} 

\paragraph{Mid-Range FPGAs}
\CL{FPGAs with more than 700 DSPs have sufficient resources for most classification or detection tasks using medium-sized models.
However, to achieve high throughput, designers must still exploit significant model optimizations, such as heavy quantization, which leads to accuracy drops.
For example, \citet{rapuanoFPGAbasedHardwareAccelerator2021a} deploy a $\sim\! 14$MB custom CNN with layer-specific QAT on an XCZU7EV and achieve inference in $141.7$ ms for $3.4$W.
Object detection tasks also induce a drop in accuracy, for example \citet{zhangFPGAImplementationCNNbased2021a} implement a lightweight version of YOLOv2, achieving $67\%$ mAP on DOTAv1.0 \cite{xiaDOTALargeScaleDataset2018}, remaining significantly below the State Of the Art (SOTA) ($82\%$). 
Similarly, \citet{zhaoHardwareAccelerationSatellite2023a} leverage PTQ, pruning, and depth-wise separable convolutions to detect objects in the DIOR \cite{liObjectDetectionOptical2020} dataset at $\sim\! 50$FPS and under $7.2$W. 
Mid-range devices can still provide low-power implementations, such as the NN HO-ShipNet \cite{ieracitanoExplainableEmbeddedNeural2024}, consuming $1.90$W on a Zynq 7045, or the ETAUS system \cite{huangEdgeTrustworthyAI2024}, consuming $1.65$W on a KV260.} 

\paragraph{High-End FPGAS}
\CL{Only a few FPGAs incorporate 3500 DSPs or more.
Such available computing power enables deployment without significant alteration of the models and facilitates results competitive with the SOTA\footnote{Comparing SOTA-driven studies with HW-aware FPGA implementations is challenging without a cost/benefit analysis.
Quantifying the computational cost of a $1\%$ accuracy gain would clarify the trade-off between task performance and hardware efficiency, yet this is often missing in SOTA literature.}. 
On the relatively old Virtex-7, \citet{niAlgorithmHardwareCoOptimization2023} deploy large models for classification and object detection tasks.
They achieve competitive accuracy on the NWPU-RESISC45 dataset \cite{chengRemoteSensingImage2017} with both ResNet-34 ($93\%$) and VGG16 ($92\%$) deployed on their \textit{Flexible} accelerator design compared to the SOTA ($96\%$).
As the DDR3 memory of the Virtex-7 can create a bottleneck on memory bandwidth, \citet{yangAlgorithmHardwareCodesign2022} stores model weights on-chip.
Through pruning and mixed-precision QAT, they consistently compress YOLOv2 with different lightweight backbones to under $0.5$MB while maintaining high accuracies. 
Their model optimizations, paired with optimized pipeline scheduling and the use of $\sim\!70\%$ of the available DSPs, lead to the lowest latencies of the surveyed DL models, achieving $0.7$ms for their MobileNetv1 \cite{howardMobileNetsEfficientConvolutional2017} and MobileNetv2 \cite{sandlerMobileNetV2InvertedResiduals2018} designs.
Two studies experiment with Alveo cards \cite{zhangAcceleratingGNNbasedSAR2023, wickramasingheVTROptimizedVision2024} and demonstrate impressive throughput performance, including for the only transformer architecture of this survey.
However, datacenter-grade boards, such as the Alveo family, have power consumptions ranging from $75$ to $225$W, making them unsuitable for edge operations.}

\paragraph{Synergizing Model and FPGA} 
\CL{Several factors influence model selection, foremost the hardware constraints discussed above and in \RQ{FPGA-motivations}.
Indeed, limited computational units (DSPs, LUTs) and on-chip storage (BRAMs) impose a trade-off between accuracy and computational throughput.
However, as seen in Fig.~\ref{fig:FPGA_metrics_subplot}, resource utilization varies significantly across all FPGA sizes.
Optimizing utilization could make smaller FPGAs viable for a broader range of applications.
Additionally, the characteristics of ML tasks and RS applications also restrict model selection.
For example, object detection problems on the edge typically rely on single-shot models, such as YOLO or SSD, to save compute time. 
For segmentation, patch-based models (e.g., UNet) tend to yield better task accuracy \cite{asgaritaghanakiDeepSemanticSegmentation2021} than methods that process pixels one at a time.
Such models, however, have a larger memory footprint and higher computational costs.
Classification, on the other hand, can be tackled with diverse models, ranging from SVMs that fit on all FPGA devices to larger models like ResNet50. 
As a final factor, because automatic toolchains only support a limited set of operations, designers who desire to accelerate development time and reduce the required design expertise must favor standard layers in their architectures.
Ultimately, optimal model selection depends on each project’s performance requirements, deployment constraints, and available development effort.}

\ifx\undefined\finalmanuscript

\paragraph{Balancing Accuracy and Computation} 
\dsstrike{Deploying large ML models on FPGAs often requires balancing task accuracy and hardware performance. 
In this regard, \citet{niAlgorithmHardwareCoOptimization2023} achieve competitive OA on the NWPU-RESISC45 dataset \cite{chengRemoteSensingImage2017} with both ResNet-34 ($93\%$) and VGG16 ($92\%$) deployed on their \textit{Flexible} accelerator design.
The authors leverage mixed-precision datatypes and multiples PEs to reach high throughput while achieving accuracies competitive with the State Of The Art (SOTA) of $96\%$.}
\dsstrike{On DOTAv1.0 \cite{xiaDOTALargeScaleDataset2018}, \citet{zhangFPGAImplementationCNNbased2021a} implement a lightweight version of YOLOv2, achieving $67\%$ mAP.
Although the accuracy remains significantly below SOTA ($82\%$), their compressed YOLOv2 model enables on-the-edge computations.} 
\CLstrike{As a general note, comparing SOTA-driven studies with HW-aware FPGA implementations is challenging without a cost/benefit analysis.
Quantifying the computational cost of a $1\%$ accuracy gain would clarify the trade-off between task performance and hardware efficiency, yet, this is often missing in SOTA literature.}

\paragraph{Compact and Fast ML Models} 
\dsstrike{Encoding models directly on-chip avoids loading parameters from peripherals, granting significant speed-up, as demonstrated by the OSCAR-RT framework in \citet{yangAlgorithmHardwareCodesign2022}.
By mapping all CNN layers onto on-chip FPGA resources and using mixed-width quantization and optimized pipeline scheduling, they consistently compress YOLOv2 with different lightweight backbones under $0.5$MB while maintaining accuracies above $90\%$ mAP.
In particular, their MobileNetv1 \cite{howardMobileNetsEfficientConvolutional2017} and MobileNetv2 \cite{sandlerMobileNetV2InvertedResiduals2018} designs run in $0.7$ms, the lowest latency of the surveyed DL models.} 
\dsstrike{\citet{weiFPGABasedHybridTypeImplementation2019} achieve LeNet-5 inference in $2.29$ms via a custom QAT algorithm, though their ablation study hints at sub-optimal FPGA resource usage.} 
\CLstrike{\citet{upadhyayDesignImplementationCNNbased2024} classify cloud coverages with their SICNet CNN in $1.8$ ms, the lowest latency of Vitis AI implementations. 
The lowest per-pixel latency in this survey is achieved by the 2D CNN implemented by \citet{heConfigurable2D3D2023a}, reaching $97 \mu s$ on a Zynq US+.
However, such performance is facilitated by the abundance of resources of the Zynq US+ series, which comes with a power consumption of $8.4$W.} 

\paragraph{Low-Power Implementations for Resource-Constrained Systems} 
\CLstrike{As \RQ{FPGA-motivations} highlights, power efficiency is the main concern for hardware designs, especially onboard SmallSats \cite{kothariFinalFrontierDeep2020}.
Traditional ML methods and shallow networks offer low power consumption, such as MLP \cite{hammoudArtificialNeuralNetworksBased2022a}, RDBC \cite{ratnakumarHighSpeedRoller2021}, Fuzzy ARTMAP \cite{yahiaouiParallelizationFuzzyARTMAP2017a}, and SVM-based models \cite{gyaneshwarRealtimeSCSUP2022} consuming less than $1$W.} 
\dsstrike{When it comes to DL, the CNN HO-ShipNet of \citet{ieracitanoExplainableEmbeddedNeural2024} consumes $1.90$W on a Zynq 7045, while the Ghost-YOLOS of \citet{yangLightweightDetectionMethod2023} uses $2.98$W.
\citet{suhAlgorithmHardwareCoOptimizationEnergyEfficient2021} achieve particularly impressive results using their custom 8-bit UniPOT quantization scheme and making use of DSP double rate, reaching $2.4$, $2.6$, and $2.0$W for their 3 SSD sizes.} 
\CLstrike{Among the solutions implemented with Vitis AI, \citet{huangEdgeTrustworthyAI2024} reports the lowest power consumption, $1.65$W, a noteworthy result for a ResNet-50-based system like their ETAUS.}

\paragraph{Lightweight Models and Quantization with Compensation} 
\CLstrike{The combination of lightweight backbones and fixed-point quantization is a recurring and effective strategy in the surveyed studies \cite{yangLightweightDetectionMethod2023, yangAlgorithmHardwareCodesign2022, yuImprovedLightweightDeep2024, liEdgeRealtimeObject2023a, liEfficientObjectDetection2019a, nerisFPGABasedImplementationCNN2022a, nguyenFPGASoCImplementationYOLOv42024, zhaoHardwareAccelerationSatellite2023a, kimOnOrbitAICloud2024}.
These two optimization methods are broadly applicable in RS, except in extreme scenarios where traditional ML methods are preferred to DNNs.
Lightweight backbone replacement or direct use of an efficient architecture, like ENet \cite{paszkeENetDeepNeural2016}, or a \textit{"tiny"} adaptation, for example, YOLOv4-tiny \cite{yuImprovedLightweightDeep2024, nguyenFPGASoCImplementationYOLOv42024}, are used in $\sim\! 20\%$ of the surveyed experiments.
To compensate for accuracy drops, designers typically enhance the baseline architecture.
For example, \citet{nguyenFPGASoCImplementationYOLOv42024} add a third YOLOv4 head to detect targets at different spatial scales and optimize training with data augmentation strategies to achieve higher baseline accuracies.
Following such enhancements, techniques like pruning or quantization can further reduce the solution's computational load.}

\fi


\begin{keymessage}{synergies}
{
\CLstrike{Accuracy loss due to design decisions during FPGA porting remains a challenge in achieving stateof-the-art results. While real-time performance is frequently achieved, most models still require
significant power ($42\%$ > $5$W). Lightweight architectures combined with moderate (> $4$-bit) quantization offer a promising trade-off between accuracy and computational efficiency.}\CL{Larger FPGAs naturally support higher model complexity and throughput, reducing the need for aggressive quantization or pruning.
While real-time performance is frequently achieved, most models still require significant power ($43\%$ $>5$W). 
High variance in resource utilization shows that efficient resource use, rather than raw device size, is a key determinant of achievable performance.}
}
\end{keymessage}

\subsection{Advancing Onboard Processing: ML Approaches for Satellites and UAVs \RQinColor{onboard}}\label{section:RQ9_onboard_AI} 
One primary motivation for this survey was to understand how AI methods currently onboard spacecraft are deployed.

\paragraph{Onboard PhiSat-1: Pioneering Work in AI for \CL{RS}\CLstrike{EO}}
PhiSat-1\footnote{$6$U CubeSat, launched 09/2020, \url{https://www.esa.int/Applications/Observing_the_Earth/Ph-sat}} is a frequently cited mission \cite{giuffridaFSat1MissionFirst2022} which deployed the first CNN model performing onboard cloud coverage classification: CloudScout \cite{giuffridaCloudScoutDeepNeural2020}.
ML-based methods accelerate data processing pipelines and recently became excellent at several essential tasks \cite{szwarcmanPrithviEO20VersatileMultiTemporal2025}.
This is particularly true in RS, where the exponentially growing amount of data has outpaced the capacity for manual expert labeling.
Deploying DL models directly onboard relieves downlink constraints and reduces the need for on-ground pre-processing. 
This saves time, conserves resources, and simplifies dataflows.
For example, CloudScout classifies the cloud coverage of freshly acquired images, discarding overly cloudy images that contain little valuable \CLstrike{EO} information.
Beyond hardware constraints, like limited memory and power, using a CNN for such a critical task implies a responsibility that comes with additional restrictions. 
As such, CloudScout has to ensure a False Positive rate under $1\%$, to avoid discarding appropriate images.
In a follow-up study, \citet{rapuanoFPGAbasedHardwareAccelerator2021a} explore the possibility of using an FPGA instead of the original Myriad 2 VPU.
They conclude that, although FPGA-based designs take longer to develop and consume more power, their superior customizability, lower latency, and radiation hardness make them suitable for long-term LEO missions.
In a recent work, \citet{cratereEfficientFPGAAcceleratedConvolutional2025} provide a comprehensive overview of the projects building on the CloudScout initiative, comparing all related cloud detection studies.

\paragraph{Onboard OPS-SAT: Experimental AI Deployment in Space}
A second ESA mission is particularly relevant to onboard ML for RS: OPS-SAT\footnote{$3$U CubeSat, 12/2019 - 05/2024, \url{https://www.esa.int/Enabling_Support/Operations/OPS-SAT}}. 
This flying laboratory featured an Altera Cyclone V SoC and was designed for experimental ML deployments.
The C-FCN++ model presented by \citet{bahlLowpowerNeuralNetworks2019a} was deployed onboard OPS-SAT.
With $273$ parameters ($47$KB), C-FCN++ is the only architecture explored in the study able to fit on-chip the Altera Cyclone V SoC.
It uses dilated convolutions to expand early receptive fields and improve performance without increasing parameter count.
Ultimately, C-FCN++ could perform cloud segmentation on a full-scale RS image in $150$ms, less than required for the camera of OPS-SAT to acquire the next image, granting real-time performance. 
For further investigations of ML onboard OPS-SAT, see \citet{kackerMachineLearningImage2022}, who tested several algorithms to prepare for the mission Beaver-Cube-2\footnote{$3$U CubeSat educational mission from MIT, expected launch 2025, \url{https://www.nanosats.eu/sat/beavercube-2}}.


\paragraph{Preparing Deep Learning for Spacecraft}
Beyond deployed solutions, several studies contribute to preparing DL models for onboard spacecraft processing.
In particular, \citet{kimOnOrbitAICloud2024} prepared for the A-HiREV mission\footnote{$6$U CubeSat platform from the Korea Aerospace Research Institute (KARI)} and present a cloud classification pipeline with goals similar to CloudScout. 
Their multi-phase approach uses three stages: a uniformity check, a lightweight ternary classifier called TriCloudNet, and a pruned U-Net.
Each stage filters out extreme cases, reducing the workload for subsequent, more computationally complex models.
This image prioritization method reduces downlink data by $40$–$50\%$ and cuts processing time and power consumption by around $50\%$.
Designed as a direct comparison to CloudScout, CloudSatNet-1 \cite{pitonakCloudSatNet1FPGABasedHardwareAccelerated2022} classifies cloud coverage to relieve downlink, with a strong focus on low-power consumption and a minimized False Positive rate.
More recently, \citet{upadhyayDesignImplementationCNNbased2024} develop SICNet, a cloud detection CNN optimized for real-time performance.
The designers focus on deployment ease using Vitis AI and prepare further applications like fire and object detection.
\citet{zhangExtremelyPipelinedFPGAbased2023a} also target downlink reduction with A2NN, a model for direct onboard classification of RS images. 
As for solutions monitoring ships, \citet{yangAlgorithmHardwareCodesign2022} present OSCAR-RT, an end-to-end algorithm/hardware codesign framework dedicated to real-time onboard ship detection with SAR images.
Their three experiments considerably focus on real-time performance using on-chip implementations.
\citet{nerisFPGABasedImplementationCNN2022a} use lightweight CNNs to identify ships and airplanes, deploying them on a Kintex US FPGA, which has a radiation-hardened counterpart, the XQRKU060---also used by \citet{rapuanoFPGAbasedHardwareAccelerator2021a}. 
\citet{ieracitanoExplainableEmbeddedNeural2024} also target ship monitoring with HO-ShipNet, a CNN enhanced with xAI techniques for increased transparency, achieving $95\%$ accuracy.
As discussed in Section~\ref{section:RQ3_FPGA_boards}, radiation resilience remains a key requirement for space deployment \cite{siegleMitigationRadiationEffects2015}.
\citet{sabogalMethodologyEvaluatingAnalyzing2021a} study SEEs effects on DL applications, quantifying performance degradation in Xilinx DPU implementations \cite{AMDVitisAI}.
Such effects are mitigated through circuit-level hardening or software solutions, like Triple Modular Redundancy (TMR).
By contrast, UAV platforms operate in less radiation-exposed environments, enabling more flexible trade-offs in model design and hardware selection. 

\paragraph{AI-Accelerated UAVs Payloads: Autonomous and Real-Time Insights}
Spacecraft are not the only edge RS platforms benefiting from ML; numerous studies explore ML-supported pipelines for UAVs.
\citet{wangAccelerationImplementationConvolutional2019} deploy a CNN on a Zynq-7000 to detect objects in images from the 2018 DAC System Design Contest, targeting real-time inference under low-power constraints.
\citet{suhAlgorithmHardwareCoOptimizationEnergyEfficient2021} design three sizes of SSD models to detect drones achieving high energy efficiency on a Zynq US+. 
Addressing a practical use case, \citet{yuImprovedLightweightDeep2024} develop an Improved YOLOv4-tiny to detect abnormal railway track fasteners.
Implemented with Vitis AI, their solution achieves $\sim\!300$ FPS and $95.1\%$ mAP, demonstrating performance levels suitable for onboard UAV deployment.
Focusing on UAV navigation, \citet{fraczekEmbeddedVisionSystem2018} experiment with Decision Trees and Support Vector Machines to classify terrain and support automated landing of UAVs.
Similarly, \citet{wangFastDetectionObstacle2024} merge optical and MilliMeter-Wave (MMW) radar data to quickly detect UAV obstacles.
Their lightweight CNN reaches real-time performance with $60$ FPS and low-power $3.3$W on a Zynq-7020.
By far the most complete surveyed UAV-focused study, \citet{huangEdgeTrustworthyAI2024} present the ETAUS system, a custom UAV based on the Pixhawk 4 drone to monitor Taiwan's Air Quality Index (AQI).
ETAUS is powered by a CNN achieving high accuracy for AQI classification and cryptographic modules implemented on the FPGA logic.
ETAUS also uses a YOLOv4 pre-trained from Vitis AI Model Zoo to detect---and later blur---private and sensitive information, such as license plates\footnote{This section of the algorithm was not detailed in the study and is therefore not reported in this survey.}.
The study focuses on maintaining high accuracy for a reliable solution and achieving sufficient FPS for real-time performance.

\paragraph{Onboard Challenges: Real-Time, Memory, and Power Constraints}
Real-time processing remains the central concern across the surveyed studies, as keeping pace with the continuous stream of newly acquired images is essential for operational success.  
To meet this demand, designers minimize model computations and maximize execution parallelism.  
In some contexts, such as the PhiSat-1 payload and CloudScout \cite{giuffridaCloudScoutDeepNeural2020}, strict memory footprint limitations introduce additional constraints.  
Low power consumption is also a recurring objective, though the diversity of mission requirements makes it difficult to extract consistent trends.  
For instance, \citet{nguyenFPGASoCImplementationYOLOv42024} could afford to onboard a YOLOv4-tiny 3L with a $26.4$W power budget on their UAV.

\begin{keymessage}{onboard}
{
Although only three studies explicitly target space missions, the widespread focus on onboard deployment suggests the broad applicability of the research to future onboard systems.
Space missions are often geared towards surveillance, while UAVs address more localized applications. 
}
\end{keymessage}

\section{The Missing Pieces: Gaps and Opportunities in FPGA-enabled ML for RS}\label{section:discussion}
This section highlights key research gaps and opportunities of this emerging field.
Following PRISMA 2020 guidelines \cite{pagePRISMA2020Explanation2021}, we first outline the limitations of our survey, then present promising study directions, and conclude with recommendations for future work.

\subsection{Survey Limitations}
While Scopus\footnote{\url{https://www.scopus.com/}} and Web of Science (WoS)\footnote{\url{https://www.webofscience.com/}} are the two most comprehensive public databases \cite{chadeganiComparisonTwoMain2013}, we limited our search to WoS to maintain focus and feasibility; a preliminary analysis showed its journal coverage better aligns with the scope of this review.
Besides, as with all systematic surveys, the search results inherently depend on the chosen query terms.
This can lead to omissions, such as works deploying on FPGAs but using broader description like 'resource-limited hardware' in their abstracts.
Similarly, although we aimed for comprehensive ML-related queries, some lesser-known models may have been overlooked.
While the heterogeneity in the surveyed literature contributes to a broad and extensive body of work, it complicates direct comparisons of experiments. 
Comparing FPGA implementations is particularly challenging, as different designs use distinct networks and devices.
Furthermore, most CNN-based studies do not benchmark against prior FPGA-based CNNs implementations, limiting our ability to identify commonly compared architectures.
Lastly, distinguishing between \textit{Flexible} and \textit{Specific} accelerators often relies on our interpretation of the described designs.

\subsection{Research Gaps Analysis} 
In this section, we identify critical research gaps from the surveyed literature.
To offer insights to guide future research directions, we structure the gaps into five subtopics. 

\paragraph{Incomplete Spectrum of RS Use Cases}
\RQnospace{applications} shows that the surveyed applications do not cover the full spectrum of RS problems. 
While edge AI does not resonate well with unstressed applications that can be processed on the ground, like long-term climate monitoring, other onboard-relevant use cases remain unexplored, such as data compression and disaster response.
Indeed, the NewSpace era enables real-time alerts for numerous emergency situations, such as wildfires.
Similarly, local authorities greatly benefit from post-disaster damage assessments, e.g., after floods \cite{mateo-garciaGlobalFloodMapping2021}. 
Exploring lightweight implementations of SOTA methods on FPGA could significantly advance both research and industry adoption.
Concurrently, DL-based data compression pipelines have gained relevance over the past decade, including for \CL{RS}\CLstrike{EO} data \cite{gomesLossyNeuralCompression2025a}.
With SmallSats imaging payloads generating data at $\approx 640$ Mbps \cite{upadhyayDesignImplementationCNNbased2024}, efficient onboard compression methods present a valuable research opportunity.
\CL{In addition to these underexplored use cases, retrieval problems remain severely underrepresented, with only two studies focusing on feature estimation, despite these tasks being frequently essential for practical remote sensing outcomes.}

\paragraph{Overlooked Prevalent DL Architectures}
Although \RQ{models} highlights the vast diversity of encountered models, the surveyed literature overlooks several widespread model families.
\CL{Indeed, almost all DL-based studies rely on CNN or GNN architectures, with a single example of transformer models \cite{wickramasingheVTROptimizedVision2024}.}\CLstrike{Indeed, all DL-based studies rely on CNN or GNN architectures, with no examples of transformers or recurrent networks.}
This gap may partly reflect the recent emergence of Vision Transformers (ViTs), which have not yet seen mature FPGA implementations.
However, it may also highlight the relative implementation simplicity of graph flows and convolution pipelines on FPGA platforms.
Nevertheless, ViTs have significantly advanced DL in RS research \cite{szwarcmanPrithviEO20VersatileMultiTemporal2025} and, despite their complexity, their superior performance makes them a promising target for edge deployment \cite{sahaVisionTransformersEdge2025a}.
Similarly, no surveyed work explores \CL{recurrent models}\CLstrike{RNNs}, despite their natural fit for temporal tasks---\CL{common}\CLstrike{recurrent} in RS, such as monitoring change across time or along flight paths.
Investigating the application of RNNs onboard flying platforms represents a valuable research direction. 
At the same time, the absence of support for transformer and recurrent architectures in FPGA toolchains (e.g., FINN, Vitis AI) poses a barrier to broader adoption.
To unlock the potential of modern DL architectures in edge environments, automated frameworks must evolve to support these emerging layers, including their quantized and pruned variants.

\paragraph{Beyond FPGAs: Coarse-Grained Reconfigurable Arrays (CGRAs)} 
\dsstrike{While \RQ{FPGA-motivations} reveals a wide diversity of FPGAs, especially high-end products suited for DL inference, we fail to find ML solutions on CGRAs.}\ds{While we discovered a wide range of FPGAs used for RS applications, we found limited use of CGRAs.}
Unlike FPGAs, CGRAs harden the commonly used arithmetic units, improving energy efficiency and allowing higher \dsstrike{unit }density.
This reduces configurability, but offers enhanced performance \ds{and power efficiency}.
As discussed in Section~\ref{section:fpga_taxonomy}, CNNs are often implemented with Processing Elements (PEs) organized in a systolic array.
The AMD Versal\footnote{\url{https://www.amd.com/en/products/adaptive-socs-and-fpgas/versal.html}} platform features a systolic array of small PEs (AI Engines) capable of basic vector operation.
\ds{\citet{perrymanDependableDPUArchitectures2025a} uses two Versal devices (AI Edge and AI Core) to implement semantic segmentation using distinct networks.
Compared to \citet{sabogalMethodologyEvaluatingAnalyzing2021a}, which uses the same U-Net model and dataset \cite{2DSemanticLabeling}, it increases the $\frac{FPS}{W}$ by 2.2x and 4.1x for the Edge and Core devices, respectively, demonstrating the potential of CGRAs.}\dsstrike{Notably, the Versal AI Edge family combines FPGA programmable logic with CGRA AI engines for on-the-edge DNNs inference.}
Given the availability of space-grade Versal devices, a detailed comparison between CGRA-based and FPGA-only solutions---particularly for highly quantized models---emerges as a concrete and promising research direction. 

\paragraph{Further Refining NNs for Efficient Inference}
\RQnospace{optimizations} reveals that some widely used compression methods in Computer Vision remain underexplored in the surveyed records.
In particular, Knowledge Distillation (KD) appears in only one study \cite{yangLightweightDetectionMethod2023}.
KD is a training strategy where a large 'teacher' model with high learning capabilities transfers its learned representations to a smaller 'student' model \cite{hintonDistillingKnowledgeNeural2015}. 
In the on-the-edge context, the student network can be tailored to the target FPGA, maximizing resource utilization and performance.
KD can also guide weight pruning to improve compression while limiting accuracy loss \cite{aghliCombiningWeightPruning2021}.
This synergy holds strong potential for deployment in constrained environments.
Although a few studies adopt hardware-aware design methods, hardware insights could benefit many optimization techniques.
Notably, no surveyed work explores mixed-precision quantization driven by layer-wise hardware feedback, as in \citet{yaoHAWQV3DyadicNeural2021}, who improve energy efficiency and latency by assigning optimal bitwidths per layer.
Similarly, we see potential in using AutoML, particularly Neural Architecture Search (NAS), to identify DNN architectures tailored to hardware constraints.
Approaches like EfficientNet \cite{tanEfficientNetRethinkingModel2020} or MobiletNetv3 \cite{howardSearchingMobileNetV32019} are designed using platform-aware NAS, but hardware-aware NAS (HW-NAS) is still a young field with significant room for growth \cite{benmezianeComprehensiveSurveyHardwareAware2021}. 
\CL{In this survey, only \citet{hammoudFPGAONBOARDPROCESSING2024} used an approach similar to HW-NAS to determine the best model for their application.}
As GPUs increasingly integrate dedicated CNN units, such techniques become even more relevant for FPGAs, which must exploit their configurability to remain competitive.
FPGA-aware NAS offers a promising path to design performant models well-matched to FPGA deployment, a research direction supported by datasets like HW-NAS-Bench \cite{liHWNASBenchHardwareAwareNeural2021} that estimate FPGA performance.

\paragraph{Trust in and Interpretability of Onboard AI} 
Another underexplored area in this survey is Uncertainty Quantification (UQ).
UQ methods estimate the confidence of model predictions to improve interpretability and support more reliable decision-making onboard.
Only \citet{myojinDetectingUncertainBNN2020} apply UQ techniques, using Monte Carlo (MC) Dropout \cite{galDropoutBayesianApproximation2016} to estimate model uncertainty through multiple stochastic forward passes.
Other established methods remain unexploited in this context and deserve further investigation; for a comprehensive overview of UQ in Deep Learning, we refer readers to \citet{gawlikowskiSurveyUncertaintyDeep2023}.
Practical benefits of UQ include uncertainty rejection, where predictions exceeding a confidence threshold are discarded, potentially easing downlink stress and increasing reliability.
Similarly, explainable AI (xAI) is crucial for critical decision-making onboard, such as alert triggering or data prioritization.
Yet, only \citet{ieracitanoExplainableEmbeddedNeural2024} employ xAI tools to interpret model decisions.
We believe that lightweight xAI and UQ methods can significantly increase the trustworthiness and adoption of AI solutions for onboard processing \cite{hohlOpeningBlackBoxSystematic2024}. 


\subsection{Recommendations for Future Work}\label{section:recommendations}

\ds{We encourage researchers to use open-source data or to publish their datasets.
When a full release is not possible due to confidentiality agreements, studies should detail, as rigorously as possible, sensor specifications, selected scenes, ground truth, data splits, etc.
Such transparency is crucial for future readers to fully comprehend the problem and replicate results.
In addition, we strongly encourage future authors to share their code and scripts as thoroughly as possible, as this provides valuable information for the readers and increases the long-term impact of their work.
Furthermore, a significant limitation across many reviewed studies is the lack of thorough metric reporting, which prevents comprehensive evaluation and comparison.
Fundamental performance indicators are often incomplete.
For example, some object detection works may report \textbf{mAP} but omit \textbf{mIoU}, preventing comparisons across studies. 
Similarly, hardware results may mention throughput and power, but use inconsistent bit-widths (e.g., \textit{i4} vs. \textit{fp32}), skewing energy efficiency comparisons. 
Even more critical, memory usage is rarely analyzed in depth.
While some papers mention model footprint, very few examine off-chip memory transfers---a common bottleneck for FPGA designs~\cite{abdelouahabAcceleratingCNNInference2018,congUnderstandingPerformanceDifferences2018}.}

\ds{Future research should also build upon existing work in FPGA implementations.
Despite the ubiquity of convolution operations in the surveyed applications, most HDL and HLS studies re-implement convolution kernels from scratch, often without an in-depth study or comparison to established designs.
This is surprising given the rich body of prior work on optimized CNN accelerators for FPGAs \cite{abdelouahabAcceleratingCNNInference2018,wangLUTNetRethinkingInference2019a,abdelfattahDLACompilerFPGA2018,umurogluFINNFrameworkFast2017}, as well as mature ASIC tape-outs from both academia \cite{chenEyerissV2Flexible2019} and industry \cite{jouppiTPUV4Optically2023}.
Rather than reinventing the convolutional microarchitecture, new studies could build on these foundations and shift their focus toward application-specific challenges.
Furthermore, we have observed that many studies highlight low FPGA resource utilization as a positive outcome; however, it may also indicate poor scalability. 
Ideally, FPGA resources should be fully utilized, except when limited by external factors like power budgets or memory bandwidth.}

\ifx\undefined\finalmanuscript

\paragraph{Remote Sensing Good Practices} 
\dsstrike{We encourage researchers to use open-source data or to publish their datasets.
When a full release is not possible due to confidentiality agreements, studies should detail, as rigorously as possible, sensor specifications, selected scenes, ground truth, data splits, etc.
Such transparency is crucial for future readers to fully comprehend the problem and replicate results.}

\paragraph{Comprehensive Reporting of Metrics} 
\dsstrike{A significant limitation across many reviewed studies is the lack of thorough metric reporting, which prevents comprehensive evaluation and comparison.
Fundamental performance indicators are often incomplete.
For example, object detection works may report \textbf{mAP}, but omit \textbf{mIoU}. 
Similarly, hardware results may mention throughput and power, but use inconsistent bit-widths (e.g., \textit{i4} vs. \textit{fp32}), skewing energy efficiency comparisons. 
Even more critical, memory usage is rarely analyzed in depth.
While some papers mention memory footprint, very few examine off-chip memory transfers---a common bottleneck for FPGA designs~\cite{abdelouahabAcceleratingCNNInference2018,congUnderstandingPerformanceDifferences2018}.
We also want to highlight the recurrent confusion around the terms \textbf{GOP}, \textbf{GOPs}, and \textbf{GOP/s}. 
We recommend using \textbf{GOP} to denote computational complexity and \textbf{GOP/s} for computational throughput. 
Similarly, the term "throughput" is often used without clarification, though it may refer to either computational throughput [\textbf{GOP/s}] or frame throughput [\textbf{FPS}].}


\paragraph{Micro-architecture of Convolution Kernels} 
\dsstrike{Despite the ubiquity of convolution operations in the surveyed applications, most HDL and HLS studies re-implement convolution kernels from scratch, often without an in-depth study or comparison to established designs.
This is surprising given the rich body of prior work on optimized CNN accelerators for FPGAs \cite{abdelouahabAcceleratingCNNInference2018,wangLUTNetRethinkingInference2019,abdelfattahDLACompilerFPGA2018,umurogluFINNFrameworkFast2017}, as well as mature ASIC tape-outs from both academia \cite{chenEyerissV2Flexible2019} and industry \cite{jouppiTenLessonsThree2021}.
Rather than reinventing the convolution micro-architecture, new studies could build on these foundations and shift focus toward application-specific challenges.
Furthermore, we have observed that many studies highlight low FPGA resource utilization as a positive outcome, however, it may also indicate poor scalability. 
Ideally, FPGA resources should be fully utilized, except when limited by external factors like power budgets or memory bandwidth.}

\paragraph{Code Availability}
\dsstrike{Among the analyzed articles, only few provide accessible code repositories.
While we recognize the difficulty of sharing runnable HDL environments, recent tools like Vitis AI \cite{AMDVitisAI}---which runs in containers and through customized Python scripts---facilitate code sharing.
We strongly encourage future authors to share their code and scripts, as they provide valuable information for the readers and increase the long-term impact of their work.} 

\fi


\section{Conclusion}\label{section:conclusion}
The integration of Machine Learning (ML) into Remote Sensing (RS) pipelines is reshaping Earth Observation, enabling real-time, autonomous data processing onboard satellites and UAVs.
Facing the growing demands of NewSpace and the need for efficient ML accelerators, FPGAs stand out as the standard payload for SmallSats due to their adaptability, potential for computational speedup, and energy efficiency.
This article followed PRISMA 2020 guidelines to systematically review the literature on ML models implemented on FPGA-based platforms for RS applications.
We proposed two taxonomies to classify existing research and answered eight research questions to map the literature landscape, unveil design paradigms, and discuss technological synergies. 
Our analysis highlighted the dominance of CNN-based architectures, the prevalence of AMD FPGAs, and the necessary synergy of quantization with FPGA programmable logic. 
Despite progress in the field, our findings reveal gaps in research, including limited diversity in RS tasks, underexplored model compression techniques, like knowledge distillation, and the scarcity of uncertainty quantification and explainability methods in FPGA-based ML.
Addressing these gaps will be crucial for advancing reliable and efficient onboard AI systems.
We hope this survey serves as a valuable resource for researchers and stimulates future advancements in FPGA-enabled Machine Learning for Remote Sensing.

\begin{acks}
This work is supported by the Helmholtz Association under the joint research school "Munich School for Data Science – MUDS”
and the German Federal Ministry of Research, Technology and Space, under the project BB-KI Chips, contract no. 16DHBKI020.
This work benefited from the use of ChatGPT-4o, which provided alternative formulations of some parts of the manuscript, for example, section titles and the abstract.
\end{acks}

\bibliographystyle{acm/ACM-Reference-Format}
\bibliography{references/all_references} 


\end{document}